\documentclass{article}

\usepackage{amssymb}
\usepackage{indentfirst} 
\usepackage{makecell}
\usepackage[table]{xcolor}
\usepackage{graphicx}
\usepackage{subcaption}
\usepackage{caption}
\usepackage{floatrow}
\newfloatcommand{capbtabbox}{table}[][\FBwidth]
\usepackage{wrapfig}
\usepackage{natbib}
\setcitestyle{numbers,square}
\usepackage{amsmath}
\usepackage{extarrows}
\usepackage{multirow}

    \usepackage[preprint]{neurips_2024}

\usepackage[utf8]{inputenc} 
\usepackage[T1]{fontenc}    
\usepackage{hyperref}       
\usepackage{url}            
\usepackage{booktabs}       
\usepackage{amsfonts}       
\usepackage{nicefrac}       
\usepackage{microtype}      

\title{BDC-Occ: Binarized Deep Convolution Unit For Binarized Occupancy Network}

\author{
	Zongkai Zhang $^{1}$, Zidong Xu $^{1}$, Wenming Yang $^{1}$\thanks{Wenming Yang is the corresponding author.}, \\ 
 \textbf{Qingmin Liao} $^1$, \textbf{Jing-Hao Xue} $^{2}$ \\
	$^{1}$ Tsinghua Shenzhen International Graduate School, Tsinghua University, China \\
    $^2$ Department of Statistical Science, University College London, UK
}

\begin{document}

\maketitle

\vspace{-7mm}
\begin{abstract}
\vspace{-3.5mm}
    Existing 3D occupancy networks demand significant hardware resources, hindering the deployment of edge devices. 
    Binarized Neural Networks (BNN) offer substantially reduced computational and memory requirements.
    However, their performance decreases notably compared to full-precision networks.
    Moreover, it is challenging to enhance the performance of binarized models by increasing the number of binarized convolutional layers, which limits their practicability for 3D occupancy prediction.
    To bridge these gaps, we propose a novel binarized deep convolution (BDC) unit that effectively enhances performance while increasing the number of binarized convolutional layers. 
    Firstly, through theoretical analysis, we demonstrate that $1\times1$ binarized convolutions introduce minimal binarization errors. 
    Therefore, additional binarized convolutional layers are constrained to $1\times1$ in the BDC unit.
    Secondly, we introduce the per-channel weight branch to mitigate the impact of binarization errors from unimportant channel features on the performance of binarized models, thereby improving performance while increasing the number of binarized convolutional layers. 
    Furthermore, we decompose the 3D occupancy network into four convolutional modules and utilize the proposed BDC unit to binarize these modules.
    Our BDC-Occ model is created by applying the proposed BDC unit to binarize the existing 3D occupancy networks.
    Comprehensive quantitative and qualitative experiments demonstrate that the proposed BDC-Occ is the state-of-the-art binarized 3D occupancy network algorithm.
    Code and models are publicly available at \url{https://github.com/zzk785089755/BDC}.
\end{abstract}

\vspace{-5mm}
\section{Introduction}
\vspace{-2.5mm}

\label{sec:intro}

Recent advancements in 3D occupancy prediction tasks have significantly impacted the fields of robotics~\cite{roboticereview, occgaussian, vastgaussian} and autonomous driving~\cite{occreview, yanxu1, radocc, detdiffusion}, emphasizing the importance of accurate perception and prediction of voxel occupancy and semantic label within 3D scenes. 
However, occupancy prediction requires predicting dense voxels, which leads to substantial computational expenses~\cite{monoscene,panoocc,occtransformer}.
Moreover, the formidable performance of occupancy prediction models relies on increasing model size~\cite{fbocc}. 
These factors collectively hinder the deployment of high-performance occupancy prediction networks on edge devices.
For instance, Convolutional Neural Networks (CNN)~\cite{resnet,alexnet,unet,fpn} possess hardware-friendly and easily deployable characteristics. 
Moreover, CNN-based occupancy prediction networks~\cite{bevdet, bevdet4d} exhibit outstanding performance, making them the primary choice for deployment on edge devices.
However, high-performance CNN-based occupancy networks~\cite{monoscene, fbocc} often involve complex computations and numerous parameters.
Therefore, it is necessary to introduce model compression techniques~\cite{com_and_acc_survey} to reduce the computational complexity and parameter count of CNN-based occupancy networks. 

Research on neural network compression and acceleration encompasses four fundamental methods: quantization~\cite{quant_survey}, pruning~\cite{prune_survey}, knowledge distillation~\cite{distillation_survey}, and lightweight network design~\cite{lightweight_survey}.
Among these methods, Binarized Neural Networks (BNN), which fall under the quantization category, quantize the weights and activations of CNN to only 1 bit, leading to significant reductions in memory and computational costs.
By quantizing both weights and activations to 1 bit, BNN~\cite{bnn} can achieve a memory compression ratio of 32$\times$ and a computational reduction of 64$\times$ when implemented on Central Processing Units (CPU). 
Furthermore, compared to full-precision models, BNN~\cite{bnn} only requires logical operations such as XNOR and bit counting, making them more easily deployable on edge devices.

Recent studies, such as BBCU~\cite{bbcu} and BiSRNet~\cite{bisrnet}, have demonstrated the capability of binarizing complex models with promising performance in tasks such as image super-resolution~\cite{irreview} and denoising~\cite{idreview}.
We try replacing each full-precision convolutional unit in the occupancy network with the binarized convolutional units proposed by these binarization algorithms.
Such a binarized model could achieve a respectable level of accuracy but still a notable performance gap compared to the full-precision model.
In full-precision models, it's common sense that increasing convolutional layers can lead to performance improvements.
However, the binarized model did not exhibit a trend of performance improvement as the number of binarized convolutional layers increased.
Instead, there is a tendency for performance to decline, making it challenging for binarized models to improve performance by increasing the number of convolutional layers~\cite{bbcu}.
Insufficient performance of binarized occupancy networks inevitably will have adverse effects on the perception of 3D space, thereby restricting the deployment of binarized models in autonomous vehicles.

\definecolor{base}{RGB}{129, 182, 223}
\definecolor{tiny}{RGB}{255, 129, 125}

\begin{figure}
  \centering
  \begin{subfigure}[b]{0.51\textwidth}
    \centering
    \includegraphics[width=\textwidth]{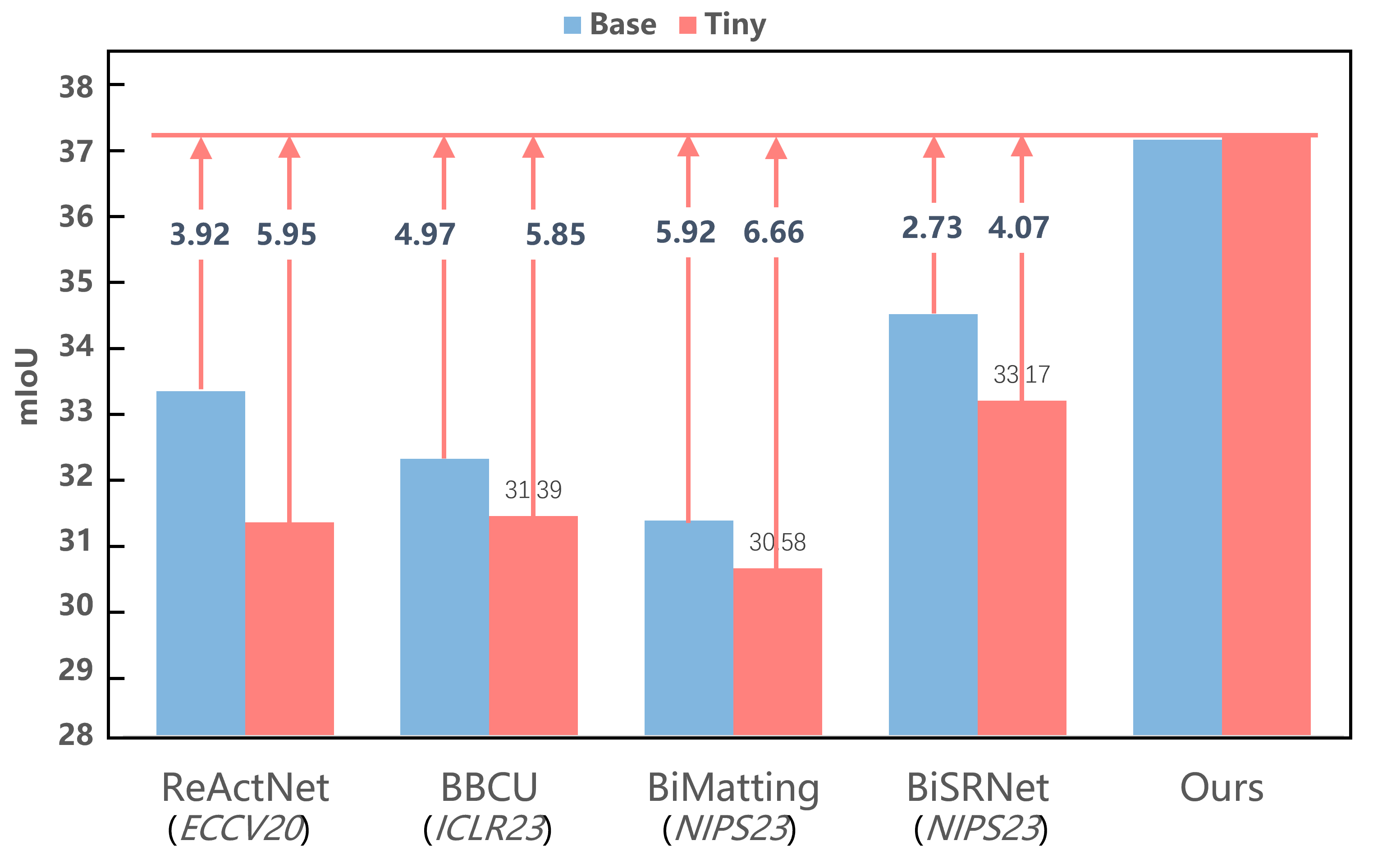}
    \caption{3D Occupancy Prediction Result}
  \end{subfigure}
  \begin{subfigure}[b]{0.48\textwidth}
    \centering
    \includegraphics[width=\textwidth]{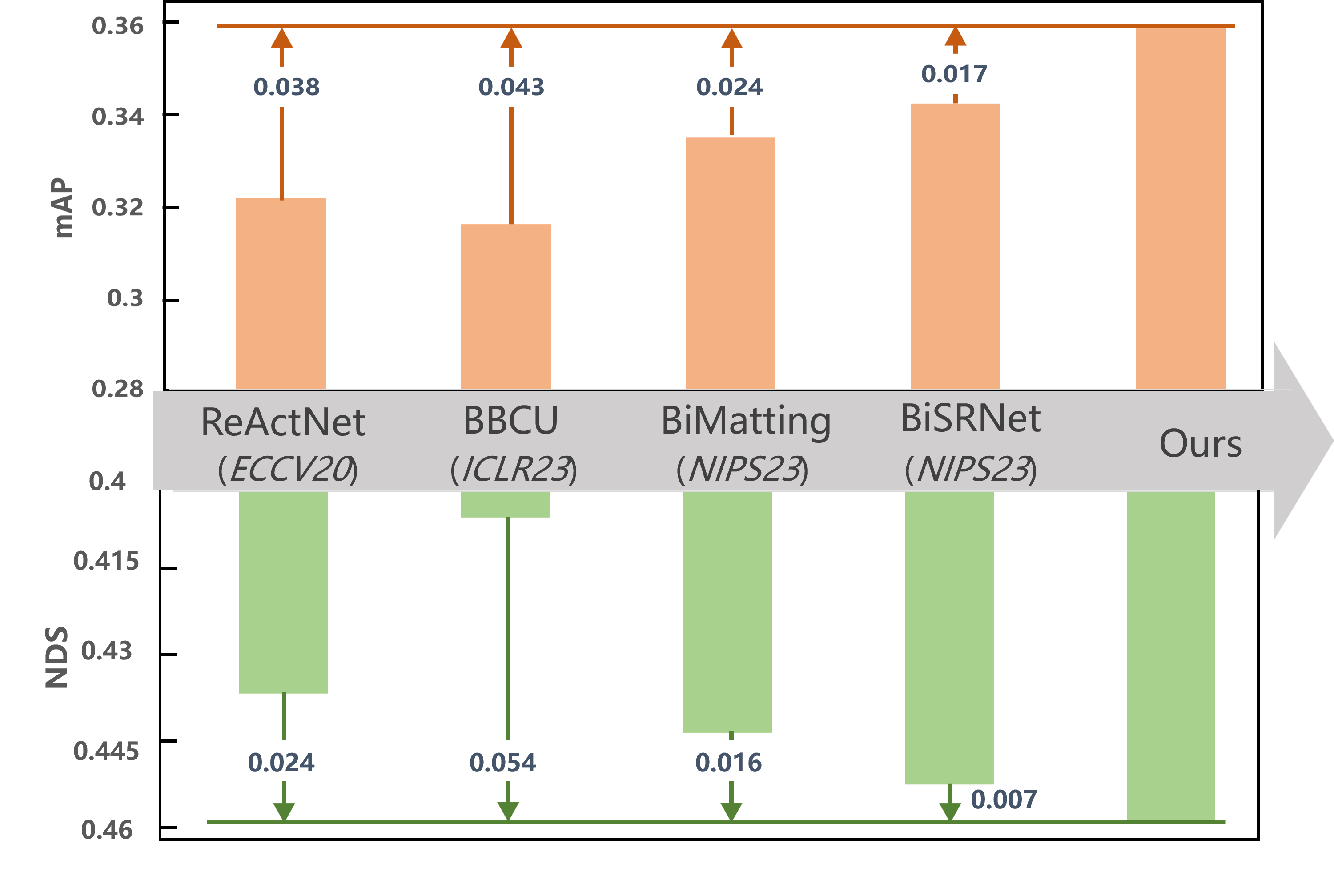}
    \caption{3D Object Detection Result}
  \end{subfigure}
  \caption{
  \textbf{Comparison between our BDC and state-of-the-art BNNs in the 3D occupancy prediction and 3D object detection tasks. }
  For the 3D occupancy prediction task, \textbf{\textcolor{base}{Base}} means binarizing the BEV encoder and occupancy head, \textbf{\textcolor{tiny}{Tiny}} means further binarizing the image neck based on \textbf{\textcolor{base}{Base}}.
  For the 3D object detection task, all binarized models are in \textbf{\textcolor{tiny}{Tiny}}.
  }
  \vspace{-3mm} 
\end{figure}

Therefore, addressing the issue of decreasing accuracy with increasing binarized convolutional layers is crucial for bridging the performance gap between binarized and full-precision models.
To address this issue, we propose a novel BNN-based method, namely Binarized Deep Convolution Occupancy (\textbf{BDC-Occ}) network for efficient and practical occupancy prediction, marking the \textbf{first} study of binarized 3D occupancy networks.
First, our theoretical analysis concludes that increasing the size of the binarized convolutional kernel also increases the binarization error.
To deepen the binarized model, we add $1 \times 1$ binarized convolutions as extra layers to reduce binarization errors.
Secondly, we propose the per-channel weight branch, placing newly added convolution layers within this branch to further enhance binarized model performance.
Thirdly, based on the above analysis, we design the \textbf{B}inarized \textbf{D}eep \textbf{C}onvolution (\textbf{BDC}) unit, which remarkably enhances binarized model performance while deepening the binarized convolution layers. 
Finally, we decompose the 3D occupancy network into four fundamental modules and customize binarization using the BDC unit for each module.

The innovations and contributions of this paper are summarized as follows:

\vspace{-1mm}
\textbf{(i)} We propose a novel BNN-based occupancy network named BDC-Occ, utilizing the \textbf{B}inarized \textbf{D}eep \textbf{C}onvolution (\textbf{BDC}) unit.
To our knowledge, it is the first work to study the binarized occupancy network.

\vspace{-1mm}
\textbf{(ii)} Based on theoretical analysis, the proposed BDC unit employs a $1 \times 1$ binarized convolution layer to deepen the binarized model.
Additionally, we introduce the per-channel weight branch to mitigate the impact of the added binarized convolution layer on binarized model performance due to binarization errors.
The 3D occupancy network is decomposed into four fundamental modules, allowing for a customized design using the BDC unit.

\vspace{-1mm}
\textbf{(iii)} We conduct extensive experiments on the Occ3D-nuScenes dataset, demonstrating significant performance improvements through deepening the binarized neural network.
Our method achieved state-of-the-art (SOTA) results, and the performance of our binarized model, BDC-Occ, is comparable to that of full-precision models.

\section{Related Work}
\vspace{-2mm}

\subsection{3D Occupancy Prediction}
The 3D occupancy prediction task comprises two sub-tasks: predicting the geometric occupancy status for each voxel in 3D space and assigning corresponding semantic labels.
We can categorize mainstream 3D occupancy networks into two architectures: CNN architecture based on the LSS~\cite{lss, simpleocc, monoscene, flashocc, sgn, inversematrixvt3d, fastocc} method and Transformer architecture based on the BEVFormer~\cite{bevformer, voxformer, tpvformer, surroundocc, symphonize, panoocc, sparseocc} method.
Due to the deployment advantages of CNN models, this paper focuses on CNN-based 3D occupancy networks.
MonoScene~\cite{monoscene} is a pioneering work that utilizes a CNN framework to extract 2D features, which it then transforms into 3D representations.
BEVDet-Occ~\cite{bevdet4d} utilizes the LSS method to convert image features into BEV (Bird's Eye View) features and employs BEV pooling techniques to accelerate model inference. 
FlashOcc~\cite{flashocc} replaces 3D convolutions in BEVDet-Occ with 2D convolutions and occupancy logits derived from 3D convolutions with channel-to-height transformations of BEV-level features obtained through 2D convolutions.
SGN~\cite{sgn} adopts a dense-sparse-dense design and proposes hybrid guidance and efficient voxel aggregation to enhance intra-class feature separation and accelerate the convergence of semantic diffusion.
InverseMatrixVT3D~\cite{inversematrixvt3d} introduces a new method based on projection matrices to construct local 3D feature volumes and global BEV features.
Despite achieving impressive results, these CNN-based methods rely on powerful hardware with substantial computational and memory resources, which are impractical for edge devices. 
How to develop 3D occupancy prediction networks for resource-constrained devices remains underexplored.
Our goal is to address this research gap.

\subsection{Binarized Neural Network}
\vspace{-2mm}

BNN~\cite{bnn, bbcu, bisrnet, bihunman, bimatting, reactnet, birealnet, xnornet, bnn-bn, irnet} represents the most extreme form of model quantization, quantizing weights and activations to just 1 bit.
Due to its significant effectiveness in memory and computational compression, BNN~\cite{bnn} finds wide application in both high-level vision and low-level vision.
For instance, Xia et al.~\cite{bbcu} designed a binarized convolutional unit, BBCU, for tasks such as image super-resolution, denoising, and reducing artifacts from JPEG compression. 
Cai et al.~\cite{bisrnet} devised a binarized convolutional unit, BiSR-Conv, capable of adjusting the density and distribution of representations for hyperspectral image (HSI) recovery. 
However, the potential of BNN in 3D occupancy tasks remains unexplored.
Hence, this paper explores binarized 3D occupancy networks, aiming to maintain high performance while minimizing computational and parameter overhead.

\section{Method}
\vspace{-2mm}

\subsection{Base Model}
\vspace{-2mm}

The full-precision models to be binarized should be lightweight and easy to deploy on edge devices.
However, prior 3D occupancy network models based on CNNs~\cite{resnet} or Transformers~\cite{transformer, swintransformer} have high computational complexity or large model sizes.
Some of these works utilize complex operations such as deformable attention, which are challenging to binarize and deploy on edge devices. 
Therefore, we redesign a simple, lightweight, and deployable baseline model without using complex computational operations.

\begin{figure}[t]
  \centering
    \includegraphics[width=1.0\linewidth]{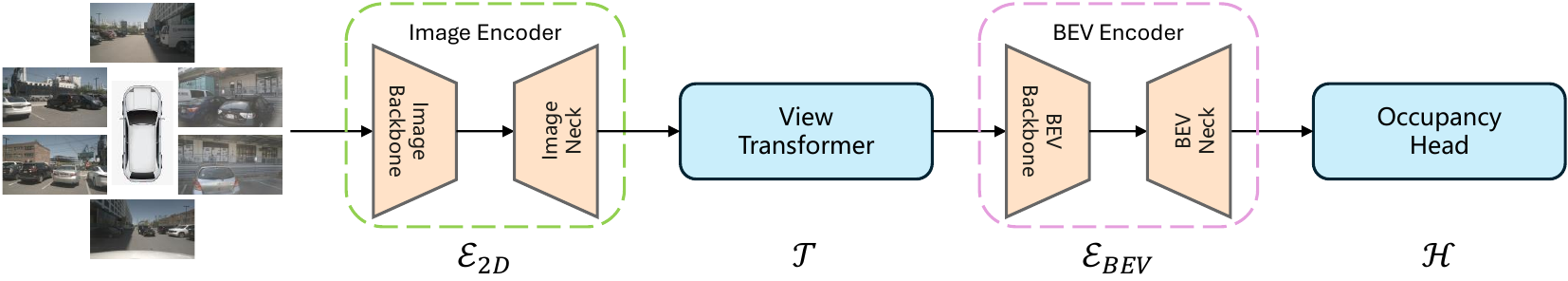}
    \caption{CNN-based 3D Occupancy Network}
  \label{fig:biocc_occnetwork}
  \vspace{-3mm} 
\end{figure}

BEVDet-Occ~\cite{bevdet} and FlashOcc~\cite{flashocc} demonstrate outstanding performance in 3D occupancy prediction tasks using only lightweight CNN architectures.
Inspired by these works, we adopt the network structure shown in Figure~\ref{fig:biocc_occnetwork} as our full-precision baseline model.
It consists of an image encoder $\mathcal{E}_{2D}$, a view transformer module $\mathcal{T}$, a BEV encoder $\mathcal{E}_{BEV}$, and an occupancy head $\mathcal{H}$. 
The occupancy prediction network is composed of these modules concatenated sequentially.
Assuming the input images are $\textbf{I} \in \mathbb{R}^{N_{view} \times 3 \times H \times W}$, the occupancy prediction output $\textbf{O} \in \mathbb{R}^{X \times Y \times Z}$ can be formulated as
\begin{equation}
  \textbf{O} = \mathcal{H}(\mathcal{E}_{BEV}(\mathcal{T}(\mathcal{E}_{2D}(\textbf{I}))))
\end{equation}
where $H$ and $W$ represent the height and width of the input images, and $X$, $Y$, and $Z$ denote the length, width, and height of the 3D space, respectively, $N_{view}$ represents the number of multi-view cameras.
Please refer to the supplementary materials for a more detailed description of the base model.

\subsection{Binarized Deep Convolution}
\vspace{-2mm}

Due to its outstanding performance and lightweight architecture, FlashOcc~\cite{flashocc} serves as the full-precision baseline model for the binarized model.
Its performance reaches 37.84 mIoU, which sets the upper performance bound for the binarized models.

Empirical evidence in full-precision models has shown that increasing network depth improves performance.
The characteristic of the binarized model allows for maintaining significantly low computational and memory usage even with increased network depth.
However, in previous research, Xia et al.~\cite{bbcu} observed that increasing the number of binarized convolutional layers within the binarized convolutional unit leads to a significant decrease in binarized model performance, the performance degradation issue with the increase in binarized convolutional layer depth within each unit restricts the further application of the binarized model. 
To address this issue, we propose the Binarized Depth Convolution (BDC) unit, which aims to enhance the binarized model performance by deepening the layers of the binarized convolution unit rather than reducing performance.

\begin{figure}[t]
  \centering
    \includegraphics[width=1\linewidth]{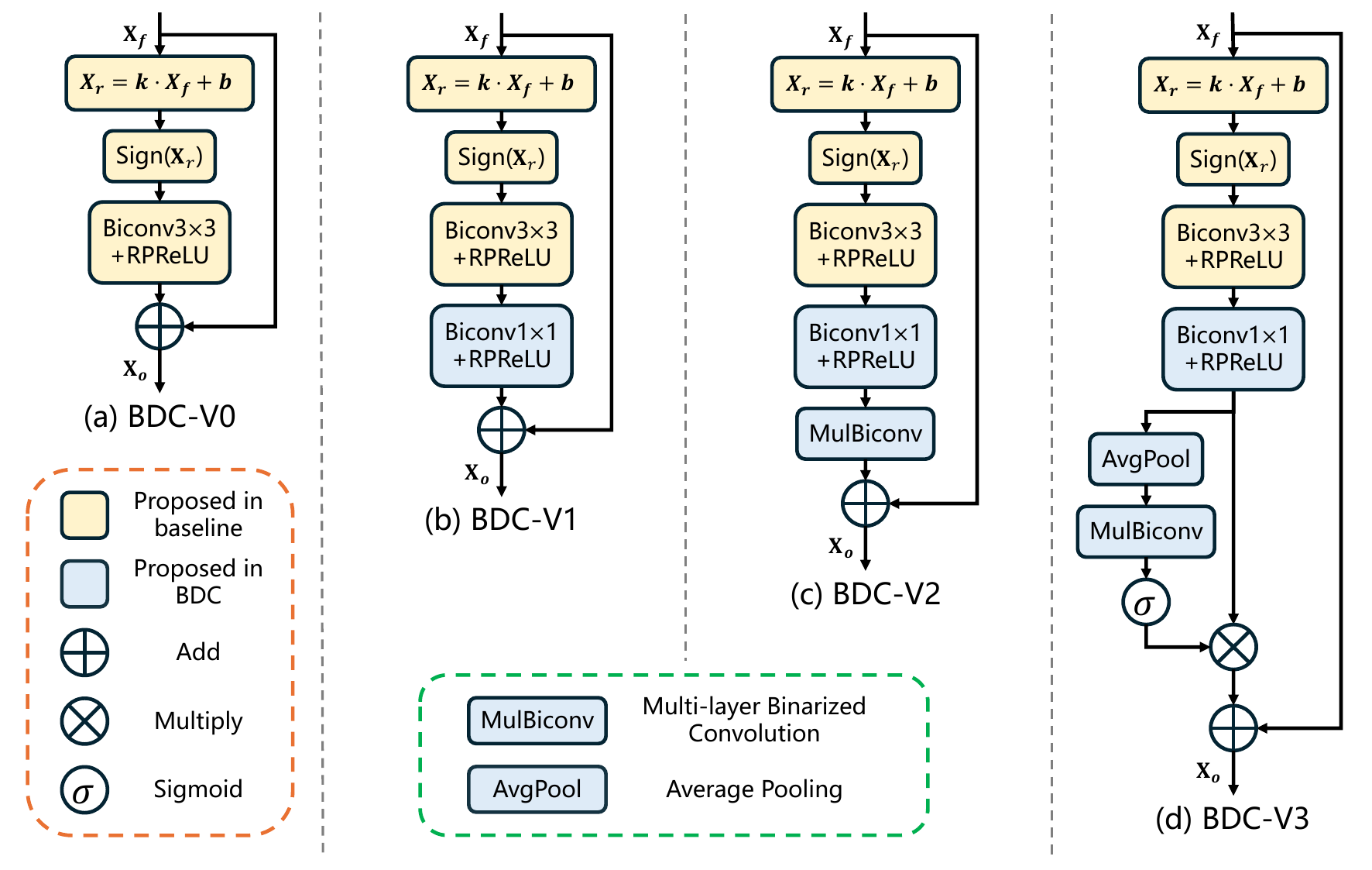}
    \caption{The illustration of the improvement process of our BDC. 
}
  \label{fig:biocc_bdc}
    \vspace{-3mm} 
\end{figure}

Cai et al.~\cite{bisrnet} proposed the binarized convolution unit BiSR-Conv, which can adjust the density and enable effective binarization of convolutional layers.
We utilize BiSR-Conv to binarize FlashOcc~\cite{flashocc}, forming our initial version of \textbf{BDC-V0}, with its structure shown in Figure~\ref{fig:biocc_bdc} (a).
Please refer to the supplementary materials for a more detailed description of the BDC-V0.
The model achieves a performance of \textbf{34.51 mIoU}.

\textbf{Theorem 1.}
\textit{
    In the process of backpropagation, we denote the expected value of the element-wise absolute gradient error of the parameters $\textbf{w}$ in the $l$-th binarized convolutional layer as $\mathbb{E}[\Delta \frac{\partial L}{\partial w_{mn}^{(l)}}]$.
    The specific expression is as follows:
  }
  \begin{equation}
       \mathbb{E}[\Delta \frac{\partial L}{\partial w_{mn}^{(l)}}] \approx 0.5354 \cdot (\sum_i \sum_j  \sum_{m'=-(k//2)}^{k//2} \sum_{n'=-(k//2)}^{k//2} \mathbb{E}[|\frac{\partial \sigma(y_{(i+m')(j+n')}^{(l)})}{\partial y_{ij}^{(l)}} \cdot  w_{m'n'}^{(l+1)} \cdot \frac{\partial L}{\partial y_{ij}^{(l+1)}}|])
    \end{equation}
\textit{
  where $k$ is the binarized convolution kernel size, $\frac{\partial \sigma(y_{(i+m')(j+n')}^{(l)})}{\partial y_{ij}^{(l)}}$ is the derivative of the activation function $\sigma(\cdot)$, $w_{m'n'}^{(l+1)}$ represents the weights of the binarized convolutional kernel in the next layer, and $\frac{\partial L}{\partial y_{ij}^{(l+1)}}$ is the element-wise gradient in the next layer.
  }

Based on Theorem 1, using a $3 \times 3$ convolutional kernel for binarized convolution leads to more binarization errors than a $1 \times 1$ kernel.
Additionally, the model necessitates the presence of the first $3 \times 3$ binarized convolution layer to maintain its capability for extracting local features. 
Therefore, building upon the binarized convolution unit BDC-V0, we introduce a $1 \times 1$ binarized convolution layer after the $3 \times 3$ binarized convolution and before the residual connection, proposing \textbf{BDC-V1} as shown in Figure \ref{fig:biocc_bdc}(b). 
By deepening the binarized convolution unit, BDC-V1 enhances its feature extraction capability, achieving a performance of \textbf{36.29 mIoU}.

To further deepen the model, we add multiple $1 \times 1$ binarized convolution layers to the binarized convolution unit named \textbf{BDC-V2}.
The structure of BDC-V2 is shown in Figure \ref{fig:biocc_bdc}(c).
We define the added multi-layer binarized convolution as $\text{MulBiconv}_N$, comprising $N$ RPReLU activations and $1\times1$ binarized convolution layers, which can be expressed as
\begin{equation}
\text{MulBiconv}_N (\cdot) = \text{Repeat}_N(\text{Biconv1}\times\text{1}(\text{RPReLU}(\cdot)))
\end{equation}
where $\text{Repeat}_N(f)$ denotes repeating $N$ times operation $f$.

When $N=1$, the performance drops to 35.88 mIoU; $N=2$, it drops further to \textbf{35.43 mIoU}.
We observe a decreasing trend in network performance as the number of $1 \times 1$ binarized convolution layers increases. 
It occurs as the accumulated binarization errors increase with the addition of more binarized convolution layers within the unit.
The negative effect on binarized model performance outweighs the positive influence of increasing parameters, resulting in decreased binarized model performance.

In BNN, not every channel feature contributes to the effectiveness of the binarized model.
Insignificant channel features carry binarization errors that contaminate important channel features, leading to performance deterioration in the binarized model.
The Squeeze-and-Excitation (SE) module proposed by SENet~\cite{senet} explicitly models interdependencies among feature channels, effectively extracting highly significant features.

Therefore, based on the idea of SENet~\cite{senet}, we propose \textbf{BDC-V3}, where we use the additional $1 \times 1$ binarized convolution to learn per-channel weights rather than features themselves.
The structure of BDC-V3 is illustrated in Figure \ref{fig:biocc_bdc}(d).
First, the output of the first $1 \times 1$ binarized convolution, $\textbf{X}_1$, serves as the input to the per-channel weight branch, which includes global average pooling (AvgPool), multi-layer binarized convolution (MulBiconv), and Sigmoid.
The branch output $\textbf{Y}_1$ is obtained by multiplying it with $\textbf{X}_1$, expressed as
\begin{equation}
  \textbf{Y}_1 = \text{Sigmoid}(\text{MulBiconv}_N (\text{AvgPool}(\textbf{X}_1))) \odot \textbf{X}_1
\end{equation}
where $\odot$ denotes element-wise multiplication. 
When $N=1$, the performance improved to 36.89 mIoU; when $N=2$, the performance further increased to \textbf{37.20 mIoU}, which is closer to 37.84 mIoU, the performance upper bound offered by the full-precision baseline model.
We chose BDC-V3 with $N=2$ as the final binarized convolutional unit, named BDConv.

\subsection{Binarized Convolution Module}
\vspace{-2mm}

Cai et al.~\cite{bisrnet} demonstrated the necessity of maintaining consistency in input and output dimensions for binarized convolutional layers to ensure the propagation of full-precision residual information.
Consequently, specialized design considerations are necessary for each binarized convolution module.
We can decompose the CNN-based occupancy network into four types of convolution modules:

(1) Basic convolution module: Input $\textbf{X} \in \mathbb{R}^{C \times H \times W}$, output $\textbf{Y} \in \mathbb{R}^{C \times H \times W}$;

(2) Down-sampling convolution module: Input $\textbf{X} \in \mathbb{R}^{C \times H \times W}$, output $\textbf{Y} \in \mathbb{R}^{2C \times \frac{H}{2} \times \frac{W}{2}}$;

(3) Up-sampling convolution module: Input $\textbf{X} \in \mathbb{R}^{C \times H \times W}$, output $\textbf{Y} \in \mathbb{R}^{C \times 2H \times 2W}$;

(4) Channel reduction convolution module: Input $\textbf{X} \in \mathbb{R}^{C \times H \times W}$, output $\textbf{Y} \in \mathbb{R}^{\frac{C}{2} \times H \times W}$;

\begin{figure}
  \centering
  \begin{subfigure}[b]{0.245\textwidth}
    \centering
    \includegraphics[width=\textwidth]{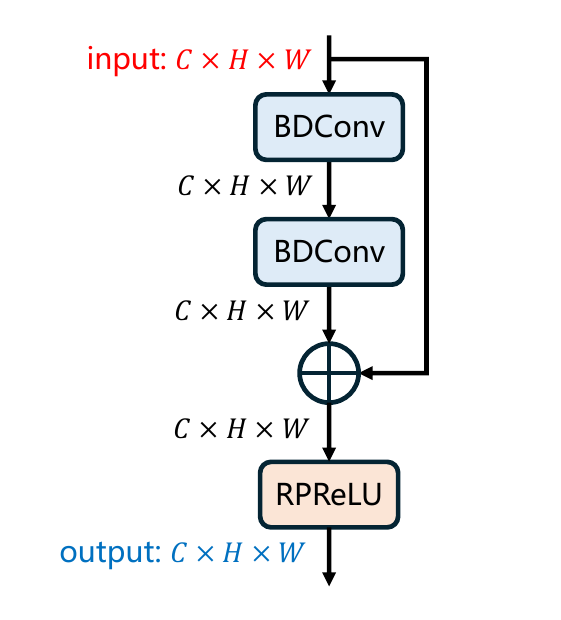}
    \caption{Basic}
    \label{fig:biocc_conv_module_a}
  \end{subfigure}
  \begin{subfigure}[b]{0.245\textwidth}
    \centering
    \includegraphics[width=\textwidth]{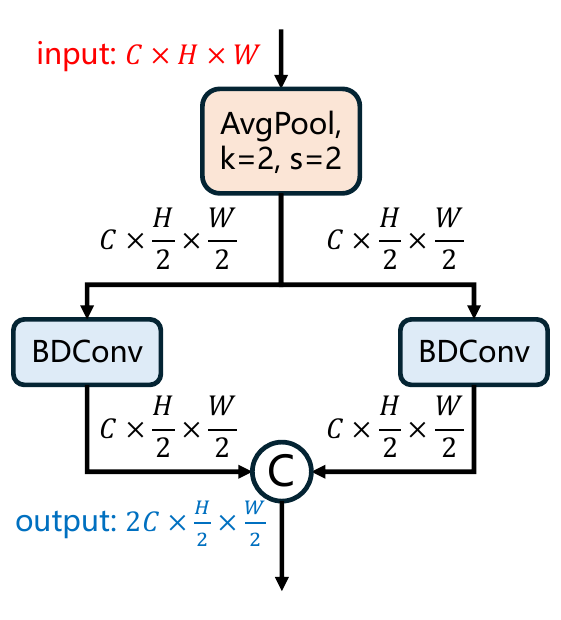}
    \caption{DownSample}
    \label{fig:biocc_conv_module_b}
  \end{subfigure}
  \begin{subfigure}[b]{0.245\textwidth}
    \centering
    \includegraphics[width=\textwidth]{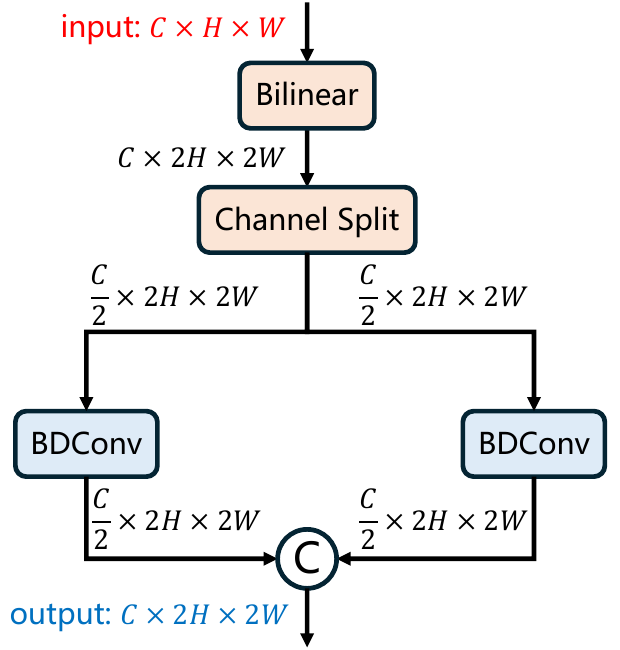}
    \caption{UpSample}
    \label{fig:biocc_conv_module_c}
  \end{subfigure}
  \begin{subfigure}[b]{0.245\textwidth}
    \centering
    \includegraphics[width=\textwidth]{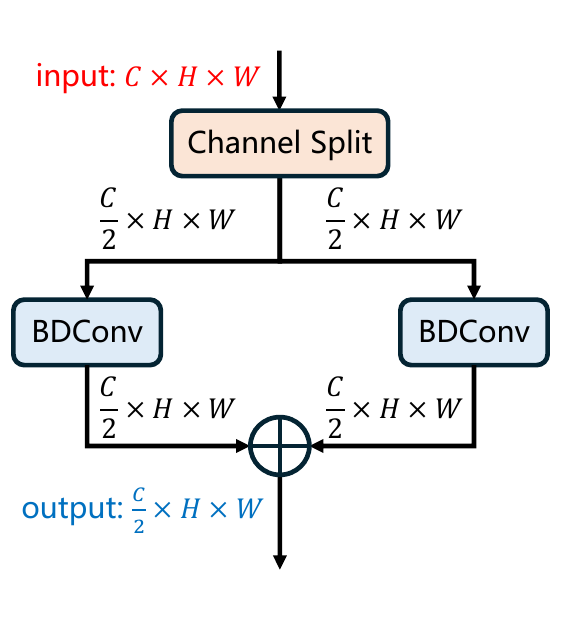}
    \caption{Channel Reduction}
    \label{fig:biocc_conv_module_d}
  \end{subfigure}
  \caption{The illustration of binarized convolution module based on BDC.}
  \label{fig:biocc_conv_module_abcd}
    \vspace{-3mm} 
\end{figure}

We adopt a binarized design approach for these four convolution modules, leveraging methodologies from previous works~\cite{reactnet, bbcu, bisrnet}, as illustrated in Figure~\ref{fig:biocc_conv_module_abcd}. 
Figure \ref{fig:biocc_conv_module_abcd} (a) illustrates the basic convolutional module, preserving both the size and the number of channels in the input feature map.
Figure~\ref{fig:biocc_conv_module_abcd} (b) depicts the downsample convolution module, reducing the size of the input feature map by half and doubling the number of channels. 
Figure~\ref{fig:biocc_conv_module_abcd} (c) showcases the upsample convolution module, doubling the size of the input feature map while preserving the number of channels. 
Finally, Figure~\ref{fig:biocc_conv_module_abcd} (d) presents the channel reduction convolution module, maintaining the size of the input feature map while halving the number of channels.

\section{Experiment}

\subsection{Experimental Settings}
\label{sec:exp_setting}

\textbf{Datasets.}
We use the Occ3D-nuScenes dataset~\cite{occ3d}, which comprises 28,130 samples for training and 6,019 samples for validation.

\textbf{Evaluation Metrics. }
We evaluate the Occ3D-nuScenes' validation set using the mean Intersection over Union (mIoU) metric.
Similar to \cite{bnn}, we compute the operations per second of BNN (OPs$^b$) as $\text{OPs}^b = \text{OPs}^f / 64$ to measure the computational complexity, where OPs$^f$ represents FLOPS.
To calculate the parameters of BNN, use the formula $\text{Parms}^b = \text{Parms}^f / 32$, where the superscript $b$ and $f$ refer to the binarized and full-precision models, respectively.
To calculate the total computational and memory costs, sum OPs as OPs$^b$ + OPs$^f$ and Params as Params$^b$ + Params$^f$.

\textbf{Implementation Details.}
For 3D occupancy prediction tasks, we employ FlashOcc~\cite{flashocc} as the baseline network.
We utilized ResNet50~\cite{resnet} as the image backbone, with an input size of \(256 \times 704\). 
Default learning rate \(1 \times 10^{-4}\), AdamW~\cite{adamw} optimizer, and weight decay of \(1 \times 10^{-2}\) were utilized. 
The training lasted approximately 29 hours, utilizing 24 epochs on two NVIDIA 3090 GPUs, with a batch size of 2 per GPU.
Data augmentation strategies for the Occ3D-nuScenes dataset remained consistent with those of FlashOcc~\cite{flashocc}.
Previous works, such as FlashOcc and BEVDet-Occ~\cite{bevdet4d}, have demonstrated the effectiveness of camera visibility masks during training.
Therefore, we also employ camera visibility masks to enhance performance.
Following the settings of FlashOcc, we employ the pre-trained model from BEVDet~\cite{bevdet} for 3D object detection tasks as our pre-training model.

\subsection{Main Results}
\vspace{-2mm}

\definecolor{nothers}{RGB}{0, 191, 255}
\definecolor{nbarrier}{RGB}{255, 120, 50}
\definecolor{nbicycle}{RGB}{255, 192, 203}
\definecolor{nbus}{RGB}{255, 255, 0}
\definecolor{ncar}{RGB}{0, 150, 245}
\definecolor{nconstruct}{RGB}{0, 255, 255}
\definecolor{nmotor}{RGB}{200, 180, 0}
\definecolor{npedestrian}{RGB}{255, 0, 0}
\definecolor{ntraffic}{RGB}{255, 240, 150}
\definecolor{ntrailer}{RGB}{135, 60, 0}
\definecolor{ntruck}{RGB}{160, 32, 240}
\definecolor{ndriveable}{RGB}{255, 0, 255}
\definecolor{nother}{RGB}{139, 137, 137}
\definecolor{nsidewalk}{RGB}{75, 0, 75}
\definecolor{nterrain}{RGB}{150, 240, 80}
\definecolor{nmanmade}{RGB}{230, 230, 250}
\definecolor{nvegetation}{RGB}{0, 175, 0}
\definecolor{lightgray}{gray}{0.9}
\definecolor{BNN}{rgb}{0.981,0.961,0.941}
\definecolor{gray}{gray}{0.85}

\definecolor{CNN}{rgb}{0.95,1,0.95}

\begin{table*}[t]

\scriptsize
 	\setlength{\tabcolsep}{0.0018\linewidth}
	
	\newcommand{\classfreq}[1]{{~\tiny(\nuscenesfreq{#1}\%)}}  %
    \begin{center}
    \caption{
    \textbf{Occupancy Prediction performance (mIoU$\uparrow$) on the Occ3D-nuScenes datasets.} Best and second best performance among BNNs are in \textcolor{red}{red} and \textcolor{blue}{blue} colors, respectively.
    }
      \vspace{-3mm} 
    \label{tab:biocc_nus}

	\begin{tabular}{l|c|c|ccccccccccccccccc|c}
		\toprule
		Methods

		& \rotatebox{90}{Params(M)} & \rotatebox{90}{OPs(G)}
  
        & \rotatebox{90}{\textcolor{nothers}{$\blacksquare$} others}
        
		& \rotatebox{90}{\textcolor{nbarrier}{$\blacksquare$} barrier}
		
		& \rotatebox{90}{\textcolor{nbicycle}{$\blacksquare$} bicycle}
		
		& \rotatebox{90}{\textcolor{nbus}{$\blacksquare$} bus}

		& \rotatebox{90}{\textcolor{ncar}{$\blacksquare$} car}

		& \rotatebox{90}{\textcolor{nconstruct}{$\blacksquare$} const. veh.}

		& \rotatebox{90}{\textcolor{nmotor}{$\blacksquare$} motorcycle}

		& \rotatebox{90}{\textcolor{npedestrian}{$\blacksquare$} pedestrian}

		& \rotatebox{90}{\textcolor{ntraffic}{$\blacksquare$} traffic cone}

		& \rotatebox{90}{\textcolor{ntrailer}{$\blacksquare$} trailer}

		& \rotatebox{90}{\textcolor{ntruck}{$\blacksquare$} truck}

		& \rotatebox{90}{\textcolor{ndriveable}{$\blacksquare$} drive. suf.}

		& \rotatebox{90}{\textcolor{nother}{$\blacksquare$} other flat}

		& \rotatebox{90}{\textcolor{nsidewalk}{$\blacksquare$} sidewalk}

		& \rotatebox{90}{\textcolor{nterrain}{$\blacksquare$} terrain}
		  
		& \rotatebox{90}{\textcolor{nmanmade}{$\blacksquare$} manmade}

		& \rotatebox{90}{\textcolor{nvegetation}{$\blacksquare$} vegetation}

        & \rotatebox{90}{mIoU}

		\\
		\midrule
        \rowcolor{CNN}\multicolumn{21}{l}{\emph{\textbf{CNN-based (32 bit)}}}\\
		\rowcolor{CNN}BEVDet-Occ~\cite{bevdet4d} & 29.02 & 241.76 & 8.22 & 44.21 & 10.34 & 42.08 & 49.63 & 23.37 & 17.41 & 21.49 & 19.70 & 31.33 & 37.09 & 80.13 & 37.37 & 50.41 & 54.29 & 45.56 & 39.59 & 36.01\\
		\rowcolor{CNN}FlashOcc~\cite{flashocc} & 44.74 & 248.57 & 9.08 & 46.32 & 17.71 & 42.70 & 50.64 & 23.72 & 20.13 & 22.34 & 24.09 & 30.26 & 37.39 & 81.68 & 40.13 & 52.34 & 56.46 & 47.69 & 40.60 & 37.84\\
        \midrule
        \rowcolor{BNN}\multicolumn{21}{l}{\emph{\textbf{BNN-based (1 bit)}}}\\
		\rowcolor{BNN}ReaActNet-T~\cite{reactnet} & 26.80 & 129.74 & 7.55 & 38.87 & 16.64 & 35.78 & 44.27 & 20.34 & 15.53 & 16.16 & 18.70 & 24.42 & 33.59 & 73.64 & 29.05 & 39.80 & 41.27 & 39.31 & 34.00 & 31.29 \\
        \rowcolor{BNN}ReaActNet-B~\cite{reactnet} & 28.17 & 133.89 & 8.62 & 40.92 & 15.94 & 37.45 & 47.23 & 18.57 & 17.47 & 18.91 & 21.52 & 23.14 & 33.13 & 77.20 & 34.58 & 45.48 & 48.31 & 42.95 & 35.06 & 33.32\\
        \midrule
		\rowcolor{BNN}BBCU-T~\cite{bbcu} & 26.79 & 129.69 & 6.24 & 38.16 & 14.33 & 31.95 & 43.18 & 20.57 & 16.50 & 17.39 & 13.45 & 22.26 & 32.51 & 75.69 & 32.97 & 42.46 & 48.50 & 41.68 & 35.75 & 31.39 \\
        \rowcolor{BNN}BBCU-B~\cite{bbcu} & 28.16 & 133.84 & 7.61 & 41.14 & 13.64 & 35.54 & 46.55 & 20.86 & 17.44 & 19.87 & 17.58 & 24.24 & 33.94 & 76.19 & 34.05 & 44.61 & 48.08 & 42.67 & 35.28 & 32.27\\
        \midrule
		\rowcolor{BNN}BiMatting-T~\cite{bimatting} & 26.82 & 129.95 & 5.96 & 38.17 & 15.27 & 35.85 & 44.11 & 19.35 & 14.38 & 18.98 & 15.84 & 23.22 & 31.16 & 73.97 & 30.51 & 35.42 & 40.9 & 41.65 & 35.05 & 30.58 \\
        \rowcolor{BNN}BiMatting-B~\cite{bimatting} & 28.17 & 134.05 & 6.80 & 38.65 & 17.99 & 33.02 & 43.80 & 19.91 & 18.29 & 18.67 & 19.82 & 21.83 & 32.09 & 72.99 & 32.44 & 41.23 & 43.64 & 36.24 & 35.07 & 31.32\\
        \midrule  
		\rowcolor{BNN}BiSRNet-T~\cite{bisrnet} & 26.79 & 129.70 & 8.38 & 41.06 & 16.76 & 33.94 & 46.11 & 18.96 & 19.10 & 17.90 & 16.94 & 23.70 & 35.14 & 76.86 & 35.68 & 46.77 & 50.39 & 41.41 & 34.78 & 33.17 \\
        \rowcolor{BNN}BiSRNet-B~\cite{bisrnet} & 28.16 & 133.85 & \textcolor{red}{\textbf{9.27}} & 41.94 & \textcolor{blue}{\textbf{19.53}} & 37.33 & 47.48 & 20.83 & \textcolor{blue}{\textbf{19.17}} & 20.08 & 20.21 & 25.36 & 33.99 & 77.42 & 35.78 & 47.35 & 50.58 & 43.24 & 37.20 & 34.51\\
        \midrule
		\rowcolor{BNN}\textbf{BDC-T (Ours)} & 26.82 & 129.90 & 9.22 & \textcolor{red}{\textbf{44.81}} & 17.56 & \textcolor{red}{\textbf{40.02}} & \textcolor{red}{\textbf{49.94}} & \textcolor{red}{\textbf{24.84}} & 18.63 & \textcolor{red}{\textbf{21.32}} & \textcolor{red}{\textbf{23.27}} & \textcolor{red}{\textbf{32.46}} & \textcolor{blue}{\textbf{36.43}} & \textcolor{red}{\textbf{81.03}} & \textcolor{blue}{\textbf{38.74}} & \textcolor{red}{\textbf{51.40}} & \textcolor{red}{\textbf{55.55}} & \textcolor{red}{\textbf{47.15}} & \textcolor{red}{\textbf{40.76}} & \textcolor{red}{\textbf{37.24}} \\
        \rowcolor{BNN}\textbf{BDC-B (Ours)} & 28.19 & 134.02 & \textcolor{blue}{\textbf{9.24}} & \textcolor{blue}{\textbf{43.93}} & \textcolor{red}{\textbf{20.25}} & \textcolor{blue}{\textbf{39.92}} & \textcolor{blue}{\textbf{49.90}} & \textcolor{blue}{\textbf{23.11}} & \textcolor{red}{\textbf{21.44}} & \textcolor{blue}{\textbf{21.22}} & \textcolor{blue}{\textbf{21.98}} & \textcolor{blue}{\textbf{31.23}} & \textcolor{red}{\textbf{37.09}} & \textcolor{blue}{\textbf{80.93}} & \textcolor{red}{\textbf{39.41}} & \textcolor{blue}{\textbf{51.03}} & \textcolor{blue}{\textbf{54.88}} & \textcolor{blue}{\textbf{46.80}} & \textcolor{blue}{\textbf{40.04}} & \textcolor{blue}{\textbf{37.20}}\\
		\bottomrule
	\end{tabular}
    \end{center}

\end{table*}

\definecolor{lightgray}{gray}{0.9}
\definecolor{BNN}{rgb}{0.981,0.961,0.941}
\definecolor{gray}{gray}{0.85}

\definecolor{CNN}{rgb}{0.95,1,0.95}

\begin{table*}[t]
  \centering
  \caption{\textbf{3D Object Detection performance (mAP$\uparrow$, NDS$\uparrow$) on the nuScenes \texttt{val} set.}
    Best performance among BNNs are in \textbf{bold}.}
  \vspace{-3mm} 
    \resizebox{\linewidth}{!}{
    \begin{tabular}{l|cc|cc|ccccc}
    \toprule

    Methods & Params(M)  &OPs(G) & \textbf{mAP}$\uparrow$ & \textbf{NDS}$\uparrow$ & mATE$\downarrow$  & mASE$\downarrow$   & mAOE$\downarrow$  & mAVE$\downarrow$  &  mAAE$\downarrow$   \\
        \midrule
    \rowcolor{CNN}\multicolumn{10}{l}{\emph{\textbf{CNN-based (32 bit)}}}\\
    \rowcolor{CNN}BEVDet~\cite{bevdet} & 44.25 & 148.77 & 0.3836 & 0.4995 & 0.5815 & 0.2790 & 0.4750 & 0.3807 & 0.2067 \\
        \midrule
    \rowcolor{BNN}\multicolumn{10}{l}{\emph{\textbf{BNN-based (1 bit)}}}\\
    \rowcolor{BNN}ReactNet-T~\cite{reactnet} & 26.53 & 101.30 & 0.3222 & 0.4358 & 0.6609 & 0.3057 & 0.6298 & 0.4468 & 0.2100  \\
    \rowcolor{BNN}BBCU-T~\cite{bbcu} & 26.51 & 101.24 & 0.3166 & 0.4046 & 0.6697 & 0.3137 & 0.7822 & 0.5461 & 0.2255  \\
    \rowcolor{BNN}BiMatting-T~\cite{bimatting} & 26.55 & 101.41 & 0.3356 & 0.4428 & \textbf{0.6358} & 0.2968 & 0.6527 & 0.4485 & 0.2159  \\
    \rowcolor{BNN}BiSRNet-T~\cite{bisrnet} & 26.52 & 101.25 & 0.3431 & 0.4519 & 0.6633 & 0.2940 & 0.5777 & 0.4550 & 0.2061 \\
    \rowcolor{BNN}BDC-T & 26.54 & 101.36 & \textbf{0.3598} & \textbf{0.4686} & 0.6362 & \textbf{0.2882} & \textbf{0.5388} & \textbf{0.4468} & \textbf{0.2030}   \\
		\bottomrule
    \end{tabular}%
    }
  \label{tab:nus_det}%
    \vspace{-3mm} 
\end{table*}%

To ensure performance, we refrain from binarizing the image backbone in the image encoder. 
This component contains pre-trained weights from image classification tasks, effectively facilitating model convergence and incorporating prior semantic information from images.
We binarize the BEV encoder and occupancy head as the \textbf{base} version (\textbf{-B}) for all binarized models.
We further binarize the image neck in the image encoder to obtain the \textbf{tiny} version (\textbf{-T}) based on the base version.

Table~\ref{tab:biocc_nus} presents the evaluation results of our method BDC on the validation set of Occ3D-nuScenes. 
To validate the effectiveness of our proposed method BDC, we compare it with other state-of-the-art binarized models, including ReActNet~\cite{reactnet}, BBCU~\cite{bbcu}, BiMatting~\cite{bihunman}, and BiSRNet~\cite{bisrnet}.
We also compare it with full-precision occupancy prediction networks based on CNN architectures, including BEVDet-Occ~\cite{bevdet} and FlashOcc~\cite{flashocc}, where FlashOcc serves as the baseline network for all binarized models and represents the theoretical upper limit of binarized model performance.

Table~\ref{tab:biocc_nus} presents performance metrics (mIoU), parameter counts, and the number of operations for different methods.
Compared to other binarized methods, our BDC-T and BDC-B achieve the best or second-best results across almost all binarized models. 
Specifically, BDC significantly improves performance without increasing parameter count or computational complexity.
Compared to the previous state-of-the-art method, BiSRNet-B, our BDC-T demonstrates superior performance in mIoU, exceeding it by \textbf{2.73 mIoU (+7.91\%)}, while saving 2.95\% of operations and 4.76\% of parameters of BiSRNet-B.
Moreover, BDC-T achieves competitive results compared to the full-precision model FlashOcc,
using only 52.26\% of operations and 59.95\% of parameters, with a minimal performance loss of \textbf{-0.6 mIoU (-1.59\%)} due to binarization errors. 
Both BBCU and BiSRNet exhibit performance degradation issues when binarizing additional modules.
Compared to BDC-B, BDC-T performs slightly better when binarizing image neck modules.
It demonstrates the robustness of BDC to the binarized modules.

To validate the generalizability of the proposed BDC, we also conduct experiments on 3D object detection tasks using the nuScenes~\cite{nuscenes} dataset.
Table~\ref{tab:nus_det} presents performance metrics for the 3D object detection task in nuScenes, where our approach, BDC, continues to demonstrate superior performance in both mAP and NDS, two crucial indicators.

\subsection{Ablation Study}

In all ablation studies, we binarize the BEV encoder and the occupancy head of the full-precision model.
This binarization setting is configured as the \textbf{base} version (\textbf{-B}) for all binarized models as described in Table~\ref{tab:biocc_nus}.

\definecolor{title}{gray}{0.95}

\begin{table*}[t]
\begin{floatrow}
\capbtabbox{
\setlength{\tabcolsep}{0.021\linewidth}
 \caption{\textbf{Break-down ablation.} Figure~\ref{fig:biocc_bdc} illustrates the structure of various versions of the BDC.} 
   \vspace{-3mm} 
     \label{tab:biocc_BDC_abl}
        \begin{tabular}{l c c c c}
            \toprule
            \rowcolor{title} Methods & mIoU & OPs (G) & Params (M) \\
            \midrule
            BDC-V0 & 34.51 & 133.85 & 28.16 \\
            BDC-V1 & 36.29 & 133.93 & 28.17 \\
            BDC-V2 & 35.43 & 134.10 & 28.19 \\
            BDC-V3 & \textbf{37.20} & 134.02 & 28.19 \\
            \bottomrule
        \end{tabular}
}{

}
\capbtabbox{
\setlength{\tabcolsep}{0.021\linewidth}
     \caption{\textbf{Kernel size ablation.} $A \rightarrow B$ represents the concatenation structure of  $A \times A$ binarized convolution followed by  $B \times B$ binarized convolution.} 
       \vspace{-3mm} 
     \label{tab:biocc_kernel_abl}
    \begin{tabular}{l c c c c }
                \toprule
                \rowcolor{title} Kernel &mIoU  &OPs (G)  &Params (M) \\
                \midrule
                $3 \rightarrow 1$ & \textbf{36.29} & 133.93 & 28.17    \\
                $3 \rightarrow 3$ & 33.01 & 133.93 & 28.17  \\
                $1 \rightarrow 1$ & 35.32 & 133.93 & 28.17    \\
                $3 \rightarrow 3 \rightarrow 1$ & 33.37 & 134.02 & 28.18   \\
                \bottomrule
    \end{tabular}
}{

 \small
}
\end{floatrow}

\end{table*}

\begin{wrapfigure}{r}{0.5\linewidth}
\small
\centering
    \caption{Ablation study of multi-layer binarized convolution (MulBiconv)
}
  \label{fig:biocc_conv_layer}
    \includegraphics[width=1\linewidth]{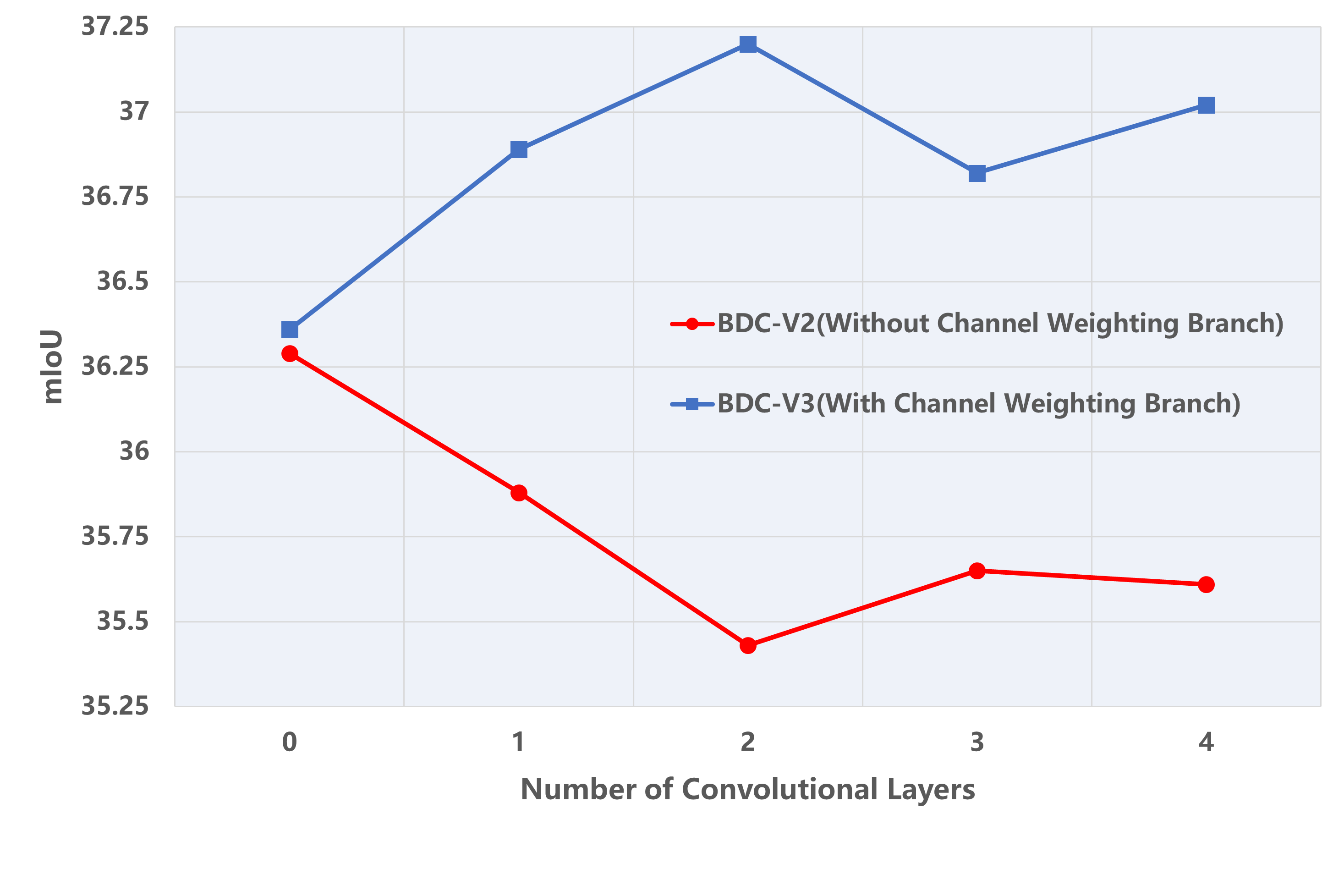}

\vspace{-5mm}
\end{wrapfigure}	

\textbf{Multi-layer Binarized Convolution (MulBiconv) Ablation.} 
To explore the impact of the number of binarized convolutional layers in MulBiconv on binarized model performance, we binarize FlashOcc using both BDC-V2 and BDC-V3 while varying the number of binarized convolutional layers in MulBiconv ($N = 0, 1, 2, 3, 4$).

The results are illustrated in Figure~\ref{fig:biocc_conv_layer}.
When $N=0$, the structure of BDC-V2 is identical to that of BDC-V1.
At this point, MulBiconv in BDC-V3 contains no learnable parameters with the per-channel weight branch.
As $N$ increases, we observe a gradual decline followed by fluctuations in the performance of BDC-V2.
In contrast, BDC-V3 initially shows performance improvement followed by fluctuations as $N$ increases.
Compared to BDC-V2, BDC-V3 exhibits a slight performance improvement.
When MulBiconv selects $N=2$, BDC-V3 achieves the best performance, reaching 37.20 mIoU.
At this point, the difference between the gain from model parameter reduction and the loss from binarization errors reaches its maximum, achieving the optimal trade-off.

\textbf{Break-down Ablation.} 
We binarize FlashOcc using four variants of BDC, where BDC-v0 represents the binarized method BiSRNet.
Additionally, for BDC-V2 and BDC-V3 utilizing the multi-layer binarized convolution (MulBiconv), we set $N=2$.

The results are presented in Table~\ref{tab:biocc_BDC_abl}, from which we can draw the following conclusions: 
(1) Compared to BDC-V0, BDC-V1 achieves a significant gain of \textbf{1.78 mIoU (+5.16\%)} by adding only one $1 \times 1$ binarized convolution layer to each binarized convolution unit. 
Due to the nature of binarized convolutions, extra binarized convolution layers result in minimal changes to model parameters and computational complexity.
(2) By adding multi-layer binarized convolution (MulBiconv) to each binarized convolution unit in BDC-V2 compared to BDC-V1, we observe a substantial decrease in performance, along with slight increases in parameter count and computational complexity.
(3) Compared to BDC-V2, BDC-V3 exhibits a significant performance improvement of \textbf{1.77 mIoU}.
Additionally, BDC-V3 gains an extra \textbf{0.91 mIoU} over BDC-V1.
Placing additional binarized convolutional layers within the per-channel weight branch effectively enhances model performance.

\textbf{Kernel Size Ablation.} 
To validate whether $3 \times 3$ binarized convolutions incur more binarization errors than $1 \times 1$ ones, potentially leading to performance degradation, we apply BDC-V1 and BDC-V2 ($N=1$) to FlashOcc.
We present the results in Table~\ref{tab:biocc_kernel_abl}.
For BDC-V1, replacing the $1 \times 1$ binarized convolution with consecutive $3 \times 3$ binarized convolutions led to a decrease in performance from 36.29 mIoU to 33.01 mIoU, with no changes in parameter count or computational complexity. 

Additionally, we validate the necessity of using a $3 \times 3$ binarized convolution as the first convolution layer. 
If replaced with a $1 \times 1$ binarized convolution, the receptive field of the binarized convolution unit becomes limited, preventing the establishment of connections with neighboring pixel features, resulting in a decrease in performance from 36.29 mIoU to 35.32 mIoU. 
Experiments conduct on BDC-V2 ($N=1$) also support the conclusion that consecutive $3 \times 3$ binarized convolutions lead to binarization errors and affect binarized model performance.

\subsection{Visualization}

\begin{figure}[t]
  \centering
    \includegraphics[width=1.0\linewidth]{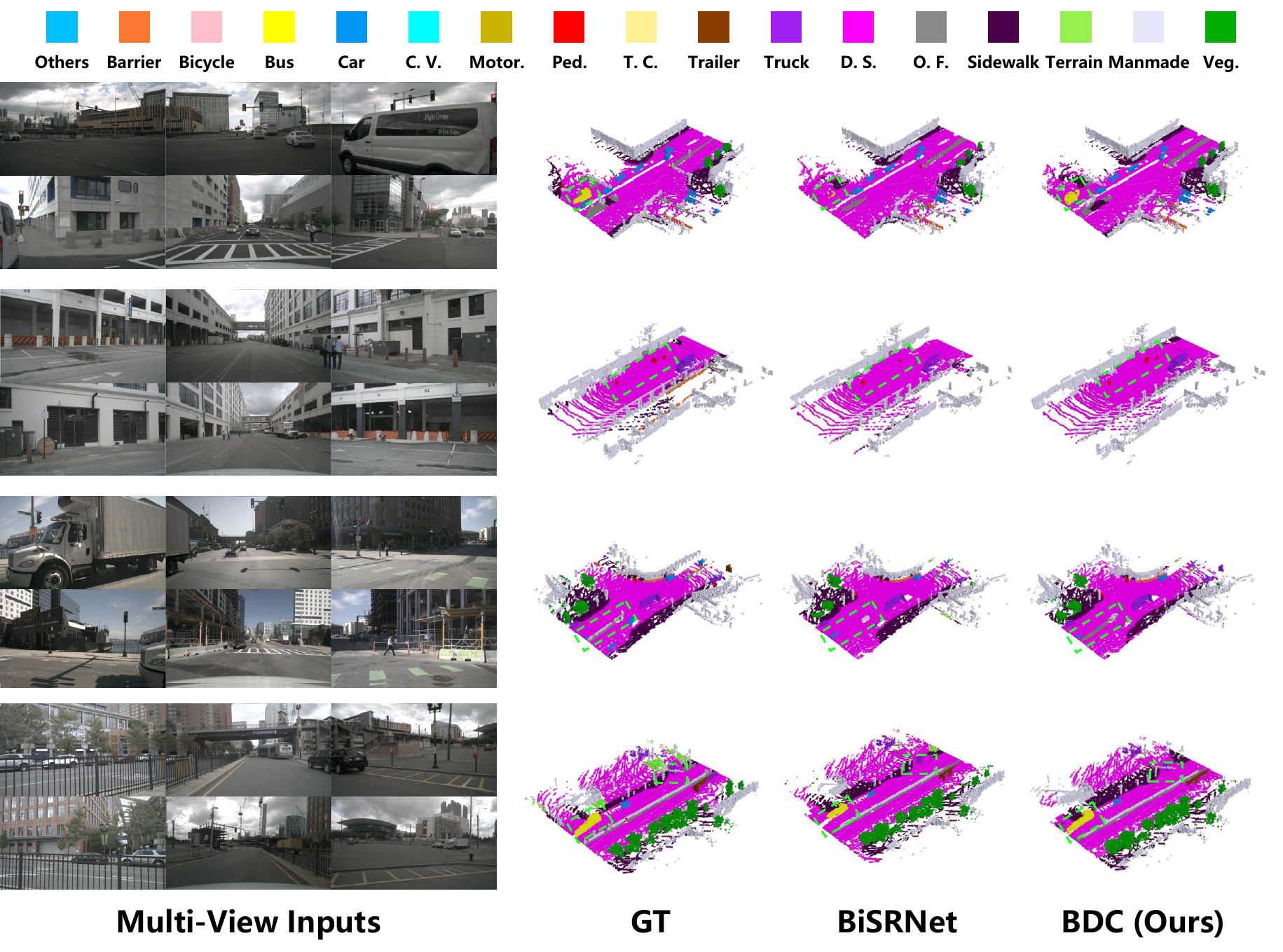}
    \caption{Visualization rensults on Occ3D-nuScenes validation set}
  \label{fig:biocc_vis}
    \vspace{-3mm} 
\end{figure}

We also present some qualitative results of BDC and BiSRNet on the Occ3D-nuScenes' validation set. 
As illustrated in Figure~\ref{fig:biocc_vis}, BDC exhibits comprehensive predictions about the bus in the first and last rows. 
In the second row, BDC successfully identifies all pedestrians, whereas BiSRNet~\cite{bisrnet} overlooks some pedestrians in the scene.
Moreover, in the third row, BDC provides accurate predictions about curbs, whereas BiSRNet misclassifies them as drivable surfaces, potentially posing safety concerns.
Additionally, in the fourth row, BDC accurately reconstructs traffic lights in the scene, showcasing its robust capability in scene perception.

\section{Conclusion}

\label{sec:conclusion}



This paper introduces a binarized deep convolution (BDC) unit for binarizing 3D occupancy networks.
The BDC unit addresses the issue observed in BNN, where increasing the number of binarized convolutional layers leads to a decrease in model performance. 
Theoretical analysis demonstrates that $1\times1$ binarized convolutions introduce minimal binarization errors during training. 
Therefore, within the BDC unit, apart from the initial binarized convolution being a $3\times3$ kernel, all others are $1\times1$ binarized convolutions.
Furthermore, BDC employs the per-channel weight branching approach to effectively mitigate the impact of binarization errors from unimportant channel features on the performance of binarized models, thereby enhancing performance while increasing the number of binarized convolutional layers.
Furthermore, extensive experiments validate the effectiveness of the BDC unit.
Our proposed method significantly outperforms existing state-of-the-art binarized convolution networks and closely approaches the performance of full-precision networks.

\textbf{Limitation.}
We have not tested our method for performance in Transformer architectures, which may limit its broader application.

\newpage

{
	\bibliographystyle{ieeetr}
	\bibliography{reference}

\begin{thebibliography}{10}

\bibitem{monoscene}
A.-Q. Cao and R.~de~Charette, ``Monoscene: Monocular 3d semantic scene completion,'' in {\em Proceedings of the IEEE/CVF Conference on Computer Vision and Pattern Recognition}, pp.~3991--4001, 2022.

\bibitem{voxformer}
Y.~Li, Z.~Yu, C.~Choy, C.~Xiao, J.~M. Alvarez, S.~Fidler, C.~Feng, and A.~Anandkumar, ``Voxformer: Sparse voxel transformer for camera-based 3d semantic scene completion,'' in {\em Proceedings of the IEEE/CVF Conference on Computer Vision and Pattern Recognition}, pp.~9087--9098, 2023.

\bibitem{tpvformer}
Y.~Huang, W.~Zheng, Y.~Zhang, J.~Zhou, and J.~Lu, ``Tri-perspective view for vision-based 3d semantic occupancy prediction,'' in {\em Proceedings of the IEEE/CVF Conference on Computer Vision and Pattern Recognition}, pp.~9223--9232, 2023.

\bibitem{occ3d}
X.~Tian, T.~Jiang, L.~Yun, Y.~Wang, Y.~Wang, and H.~Zhao, ``Occ3d: A large-scale 3d occupancy prediction benchmark for autonomous driving,'' {\em arXiv preprint arXiv:2304.14365}, 2023.

\bibitem{bevdet}
J.~Huang, G.~Huang, Z.~Zhu, Y.~Ye, and D.~Du, ``Bevdet: High-performance multi-camera 3d object detection in bird-eye-view,'' {\em arXiv preprint arXiv:2112.11790}, 2021.

\bibitem{bevdet4d}
J.~Huang and G.~Huang, ``Bevdet4d: Exploit temporal cues in multi-camera 3d object detection,'' {\em arXiv preprint arXiv:2203.17054}, 2022.

\bibitem{bevformer}
Z.~Li, W.~Wang, H.~Li, E.~Xie, C.~Sima, T.~Lu, Y.~Qiao, and J.~Dai, ``Bevformer: Learning bird’s-eye-view representation from multi-camera images via spatiotemporal transformers,'' in {\em European conference on computer vision}, pp.~1--18, Springer, 2022.

\bibitem{lss}
J.~Philion and S.~Fidler, ``Lift, splat, shoot: Encoding images from arbitrary camera rigs by implicitly unprojecting to 3d,'' in {\em Computer Vision--ECCV 2020: 16th European Conference, Glasgow, UK, August 23--28, 2020, Proceedings, Part XIV 16}, pp.~194--210, Springer, 2020.

\bibitem{resnet}
K.~He, X.~Zhang, S.~Ren, and J.~Sun, ``Deep residual learning for image recognition,'' in {\em Proceedings of the IEEE conference on computer vision and pattern recognition}, pp.~770--778, 2016.

\bibitem{fbocc}
Z.~Li, Z.~Yu, D.~Austin, M.~Fang, S.~Lan, J.~Kautz, and J.~M. Alvarez, ``Fb-occ: 3d occupancy prediction based on forward-backward view transformation,'' {\em arXiv preprint arXiv:2307.01492}, 2023.

\bibitem{nuscenes}
H.~Caesar, V.~Bankiti, A.~H. Lang, S.~Vora, V.~E. Liong, Q.~Xu, A.~Krishnan, Y.~Pan, G.~Baldan, and O.~Beijbom, ``nuscenes: A multimodal dataset for autonomous driving,'' in {\em Proceedings of the IEEE/CVF conference on computer vision and pattern recognition}, pp.~11621--11631, 2020.

\bibitem{cityscapes}
M.~Cordts, M.~Omran, S.~Ramos, T.~Rehfeld, M.~Enzweiler, R.~Benenson, U.~Franke, S.~Roth, and B.~Schiele, ``The cityscapes dataset for semantic urban scene understanding,'' in {\em Proceedings of the IEEE conference on computer vision and pattern recognition}, pp.~3213--3223, 2016.

\bibitem{swintransformer}
Z.~Liu, Y.~Lin, Y.~Cao, H.~Hu, Y.~Wei, Z.~Zhang, S.~Lin, and B.~Guo, ``Swin transformer: Hierarchical vision transformer using shifted windows,'' in {\em Proceedings of the IEEE/CVF international conference on computer vision}, pp.~10012--10022, 2021.

\bibitem{transformer}
A.~Dosovitskiy, L.~Beyer, A.~Kolesnikov, D.~Weissenborn, X.~Zhai, T.~Unterthiner, M.~Dehghani, M.~Minderer, G.~Heigold, S.~Gelly, {\em et~al.}, ``An image is worth 16x16 words: Transformers for image recognition at scale,'' {\em arXiv preprint arXiv:2010.11929}, 2020.

\bibitem{semantickitti}
J.~Behley, M.~Garbade, A.~Milioto, J.~Quenzel, S.~Behnke, C.~Stachniss, and J.~Gall, ``Semantickitti: A dataset for semantic scene understanding of lidar sequences,'' in {\em Proceedings of the IEEE/CVF international conference on computer vision}, pp.~9297--9307, 2019.

\bibitem{kitti}
A.~Geiger, P.~Lenz, C.~Stiller, and R.~Urtasun, ``Vision meets robotics: The kitti dataset,'' {\em The International Journal of Robotics Research}, vol.~32, no.~11, pp.~1231--1237, 2013.

\bibitem{inversematrixvt3d}
Z.~Ming, J.~S. Berrio, M.~Shan, and S.~Worrall, ``Inversematrixvt3d: An efficient projection matrix-based approach for 3d occupancy prediction,'' {\em arXiv preprint arXiv:2401.12422}, 2024.

\bibitem{bisrnet}
Y.~Cai, Y.~Zheng, J.~Lin, X.~Yuan, Y.~Zhang, and H.~Wang, ``Binarized spectral compressive imaging,'' {\em Advances in Neural Information Processing Systems}, vol.~36, 2024.

\bibitem{bnn}
I.~Hubara, M.~Courbariaux, D.~Soudry, R.~El-Yaniv, and Y.~Bengio, ``Binarized neural networks,'' {\em Advances in neural information processing systems}, vol.~29, 2016.

\bibitem{xnornet}
M.~Rastegari, V.~Ordonez, J.~Redmon, and A.~Farhadi, ``Xnor-net: Imagenet classification using binary convolutional neural networks,'' in {\em European conference on computer vision}, pp.~525--542, Springer, 2016.

\bibitem{birealnet}
Z.~Liu, B.~Wu, W.~Luo, X.~Yang, W.~Liu, and K.-T. Cheng, ``Bi-real net: Enhancing the performance of 1-bit cnns with improved representational capability and advanced training algorithm,'' in {\em Proceedings of the European conference on computer vision (ECCV)}, pp.~722--737, 2018.

\bibitem{irnet}
H.~Qin, R.~Gong, X.~Liu, M.~Shen, Z.~Wei, F.~Yu, and J.~Song, ``Forward and backward information retention for accurate binary neural networks,'' in {\em Proceedings of the IEEE/CVF conference on computer vision and pattern recognition}, pp.~2250--2259, 2020.

\bibitem{reactnet}
Z.~Liu, Z.~Shen, M.~Savvides, and K.-T. Cheng, ``Reactnet: Towards precise binary neural network with generalized activation functions,'' in {\em Computer Vision--ECCV 2020: 16th European Conference, Glasgow, UK, August 23--28, 2020, Proceedings, Part XIV 16}, pp.~143--159, Springer, 2020.

\bibitem{bbcu}
B.~Xia, Y.~Zhang, Y.~Wang, Y.~Tian, W.~Yang, R.~Timofte, and L.~Van~Gool, ``Basic binary convolution unit for binarized image restoration network,'' {\em arXiv preprint arXiv:2210.00405}, 2022.

\bibitem{flashocc}
Z.~Yu, C.~Shu, J.~Deng, K.~Lu, Z.~Liu, J.~Yu, D.~Yang, H.~Li, and Y.~Chen, ``Flashocc: Fast and memory-efficient occupancy prediction via channel-to-height plugin,'' {\em arXiv preprint arXiv:2311.12058}, 2023.

\bibitem{senet}
J.~Hu, L.~Shen, and G.~Sun, ``Squeeze-and-excitation networks,'' in {\em Proceedings of the IEEE conference on computer vision and pattern recognition}, pp.~7132--7141, 2018.

\bibitem{bimatting}
H.~Qin, L.~Ke, X.~Ma, M.~Danelljan, Y.-W. Tai, C.-K. Tang, X.~Liu, and F.~Yu, ``Bimatting: Efficient video matting via binarization,'' {\em Advances in Neural Information Processing Systems}, vol.~36, 2024.

\bibitem{bihunman}
Z.~Li, Y.~Zhang, J.~Lin, H.~Qin, J.~Gu, X.~Yuan, L.~Kong, and X.~Yang, ``Binarized 3d whole-body human mesh recovery,'' {\em arXiv preprint arXiv:2311.14323}, 2023.

\bibitem{bnn-bn}
T.~Chen, Z.~Zhang, X.~Ouyang, Z.~Liu, Z.~Shen, and Z.~Wang, ``" bnn-bn=?": Training binary neural networks without batch normalization,'' in {\em Proceedings of the IEEE/CVF conference on computer vision and pattern recognition}, pp.~4619--4629, 2021.

\bibitem{sgn}
J.~Mei, Y.~Yang, M.~Wang, J.~Zhu, X.~Zhao, J.~Ra, L.~Li, and Y.~Liu, ``Camera-based 3d semantic scene completion with sparse guidance network,'' {\em arXiv preprint arXiv:2312.05752}, 2023.

\bibitem{panoocc}
Y.~Wang, Y.~Chen, X.~Liao, L.~Fan, and Z.~Zhang, ``Panoocc: Unified occupancy representation for camera-based 3d panoptic segmentation,'' {\em arXiv preprint arXiv:2306.10013}, 2023.

\bibitem{occtransformer}
J.~Liu, S.~Zhang, C.~Kong, W.~Zhang, Y.~Wu, Y.~Ding, B.~Xu, R.~Ming, D.~Wei, and X.~Liu, ``Occtransformer: Improving bevformer for 3d camera-only occupancy prediction,'' {\em arXiv preprint arXiv:2402.18140}, 2024.

\bibitem{alexnet}
A.~Krizhevsky, I.~Sutskever, and G.~E. Hinton, ``Imagenet classification with deep convolutional neural networks,'' {\em Communications of the ACM}, vol.~60, no.~6, pp.~84--90, 2017.

\bibitem{unet}
O.~Ronneberger, P.~Fischer, and T.~Brox, ``U-net: Convolutional networks for biomedical image segmentation,'' in {\em Medical image computing and computer-assisted intervention--MICCAI 2015: 18th international conference, Munich, Germany, October 5-9, 2015, proceedings, part III 18}, pp.~234--241, Springer, 2015.

\bibitem{fpn}
T.-Y. Lin, P.~Doll{\'a}r, R.~Girshick, K.~He, B.~Hariharan, and S.~Belongie, ``Feature pyramid networks for object detection,'' in {\em Proceedings of the IEEE conference on computer vision and pattern recognition}, pp.~2117--2125, 2017.

\bibitem{occreview}
Y.~Shi, K.~Jiang, J.~Li, J.~Wen, Z.~Qian, M.~Yang, K.~Wang, and D.~Yang, ``Grid-centric traffic scenario perception for autonomous driving: A comprehensive review,'' {\em arXiv preprint arXiv:2303.01212}, 2023.

\bibitem{roboticereview}
G.~N. DeSouza and A.~C. Kak, ``Vision for mobile robot navigation: A survey,'' {\em IEEE transactions on pattern analysis and machine intelligence}, vol.~24, no.~2, pp.~237--267, 2002.

\bibitem{irreview}
W.~Yang, X.~Zhang, Y.~Tian, W.~Wang, J.-H. Xue, and Q.~Liao, ``Deep learning for single image super-resolution: A brief review,'' {\em IEEE Transactions on Multimedia}, vol.~21, no.~12, pp.~3106--3121, 2019.

\bibitem{idreview}
C.~Tian, L.~Fei, W.~Zheng, Y.~Xu, W.~Zuo, and C.-W. Lin, ``Deep learning on image denoising: An overview,'' {\em Neural Networks}, vol.~131, pp.~251--275, 2020.

\bibitem{adamw}
I.~Loshchilov and F.~Hutter, ``Decoupled weight decay regularization,'' {\em Learning,Learning}, Nov 2017.

\bibitem{surroundocc}
Y.~Wei, L.~Zhao, W.~Zheng, Z.~Zhu, J.~Zhou, and J.~Lu, ``Surroundocc: Multi-camera 3d occupancy prediction for autonomous driving,'' in {\em Proceedings of the IEEE/CVF International Conference on Computer Vision}, pp.~21729--21740, 2023.

\bibitem{symphonize}
H.~Jiang, T.~Cheng, N.~Gao, H.~Zhang, W.~Liu, and X.~Wang, ``Symphonize 3d semantic scene completion with contextual instance queries,'' {\em arXiv preprint arXiv:2306.15670}, 2023.

\bibitem{simpleocc}
W.~Gan, N.~Mo, H.~Xu, and N.~Yokoya, ``A simple attempt for 3d occupancy estimation in autonomous driving,'' {\em arXiv preprint arXiv:2303.10076}, 2023.

\bibitem{sparseocc}
H.~Liu, H.~Wang, Y.~Chen, Z.~Yang, J.~Zeng, L.~Chen, and L.~Wang, ``Fully sparse 3d panoptic occupancy prediction,'' {\em arXiv preprint arXiv:2312.17118}, 2023.

\bibitem{ucc}
J.~Fan, B.~Gao, H.~Jin, and L.~Jiang, ``Ucc: Uncertainty guided cross-head co-training for semi-supervised semantic segmentation,'' in {\em Proceedings of the IEEE/CVF conference on computer vision and pattern recognition}, pp.~9947--9956, 2022.

\bibitem{detdiffusion}
Y.~Wang, R.~Gao, K.~Chen, K.~Zhou, Y.~Cai, L.~Hong, Z.~Li, L.~Jiang, D.-Y. Yeung, Q.~Xu, {\em et~al.}, ``Detdiffusion: Synergizing generative and perceptive models for enhanced data generation and perception,'' {\em arXiv preprint arXiv:2403.13304}, 2024.

\bibitem{yanxu1}
X.~Yan, H.~Zhang, Y.~Cai, J.~Guo, W.~Qiu, B.~Gao, K.~Zhou, Y.~Zhao, H.~Jin, J.~Gao, {\em et~al.}, ``Forging vision foundation models for autonomous driving: Challenges, methodologies, and opportunities,'' {\em arXiv preprint arXiv:2401.08045}, 2024.

\bibitem{radocc}
H.~Zhang, X.~Yan, D.~Bai, J.~Gao, P.~Wang, B.~Liu, S.~Cui, and Z.~Li, ``Radocc: Learning cross-modality occupancy knowledge through rendering assisted distillation,'' in {\em Proceedings of the AAAI Conference on Artificial Intelligence}, vol.~38, pp.~7060--7068, 2024.

\bibitem{occgaussian}
J.~Ye, Z.~Zhang, Y.~Jiang, Q.~Liao, W.~Yang, and Z.~Lu, ``Occgaussian: 3d gaussian splatting for occluded human rendering,'' {\em arXiv preprint arXiv:2404.08449}, 2024.

\bibitem{vastgaussian}
J.~Lin, Z.~Li, X.~Tang, J.~Liu, S.~Liu, J.~Liu, Y.~Lu, X.~Wu, S.~Xu, Y.~Yan, {\em et~al.}, ``Vastgaussian: Vast 3d gaussians for large scene reconstruction,'' {\em arXiv preprint arXiv:2402.17427}, 2024.

\bibitem{com_and_acc_survey}
L.~Deng, G.~Li, S.~Han, L.~Shi, and Y.~Xie, ``Model compression and hardware acceleration for neural networks: A comprehensive survey,'' {\em Proceedings of the IEEE}, vol.~108, no.~4, pp.~485--532, 2020.

\bibitem{quant_survey}
A.~Gholami, S.~Kim, Z.~Dong, Z.~Yao, M.~W. Mahoney, and K.~Keutzer, ``A survey of quantization methods for efficient neural network inference,'' in {\em Low-Power Computer Vision}, pp.~291--326, Chapman and Hall/CRC, 2022.

\bibitem{prune_survey}
T.~Liang, J.~Glossner, L.~Wang, S.~Shi, and X.~Zhang, ``Pruning and quantization for deep neural network acceleration: A survey,'' {\em Neurocomputing}, vol.~461, pp.~370--403, 2021.

\bibitem{distillation_survey}
J.~Gou, B.~Yu, S.~J. Maybank, and D.~Tao, ``Knowledge distillation: A survey,'' {\em International Journal of Computer Vision}, vol.~129, no.~6, pp.~1789--1819, 2021.

\bibitem{lightweight_survey}
Y.~Zhou, S.~Chen, Y.~Wang, and W.~Huan, ``Review of research on lightweight convolutional neural networks,'' in {\em 2020 IEEE 5th Information Technology and Mechatronics Engineering Conference (ITOEC)}, pp.~1713--1720, IEEE, 2020.

\bibitem{fastocc}
J.~Hou, X.~Li, W.~Guan, G.~Zhang, D.~Feng, Y.~Du, X.~Xue, and J.~Pu, ``Fastocc: Accelerating 3d occupancy prediction by fusing the 2d bird's-eye view and perspective view,'' {\em arXiv preprint arXiv:2403.02710}, 2024.

\end{thebibliography}
}

\newpage

\section{Appendix}
\label{sec:appendix}

\subsection{More Details About Base Model}

Base model consists of an image encoder $\mathcal{E}_{2D}$, a view transformer module $\mathcal{T}$, a BEV encoder $\mathcal{E}_{BEV}$, and an occupancy head $\mathcal{H}$. 
The occupancy prediction network is composed of these modules concatenated sequentially.
Assuming the input images are $\textbf{I} \in \mathbb{R}^{N_{view} \times 3 \times H \times W}$, the occupancy prediction output $\textbf{O} \in \mathbb{R}^{X \times Y \times Z}$ can be formulated as
\begin{equation}
  \textbf{O} = \mathcal{H}(\mathcal{E}_{BEV}(\mathcal{T}(\mathcal{E}_{2D}(\textbf{I}))))
\end{equation}
where $H$ and $W$ represent the height and width of the input images, and $X$, $Y$, and $Z$ denote the length, width, and height of the 3D space, respectively, $N_{view}$ represents the number of multi-view cameras.

First, Multi-view images are sent to the image encoder $\mathcal{E}_{2D}$ to obtain 2D features $\textbf{f}_{2D} \in \mathbb{R}^{N_{\text{view}} \times C_{2D} \times H_{2D} \times W_{2D}}$ and depth prediction $\textbf{f}_{\text{depth}} \in \mathbb{R}^{N_{\text{view}} \times N_{\text{depth}} \times H_{2D} \times W_{2D}}$, where $C_{2D}, H_{2D}, W_{2D}$ denote the number of channels, height and width of 2D features, respectively. 
$N_{depth}$ represents the number of depth bins in the depth prediction.

Subsequently, the image features $\textbf{f}_{2D}$ and depth prediction $\textbf{f}_{depth}$ are passed through the visual transformation module $\mathcal{T}$, which transforms them into primary BEV features $\textbf{f}_T \in \mathbb{R}^{C_{BEV} \times H_{BEV} \times W_{BEV}}$ using camera intrinsic and extrinsic projection matrices. 
Here, $C_{BEV}$ represents the number of channels of BEV features, while $H_{BEV}$ and $W_{BEV}$ represent the length and width of the BEV space, respectively. 
Since the voxel distribution obtained from the depth map through projection matrices is sparse, the representation capability of primary BEV features may be insufficient.
To this end, $\textbf{f}_T$ is passed through the BEV encoder $\mathcal{E}_{BEV3D}$ to obtain fine BEV features $\textbf{f}_{BEV} \in \mathbb{R}^{C_{BEV} \times H_{BEV} \times W_{BEV}}$ for further refinement.

Finally, the semantic prediction output logits $\textbf{O}_{logits} \in \mathbb{R}^{N_{class} \times X \times Y \times Z}$ come from the BEV features $\textbf{f}_{BEV}$ processed through the occupancy prediction head $\mathcal{H}$, where $N_{class}$ is the number of semantic classes in the dataset.
By taking the index corresponding to the maximum value of the logits, we can obtain the final occupancy prediction output $\textbf{O}$.

\subsection{More Details About BDC-V0}

We define BDC-V0 following the method proposed in BiSRNet~\cite{bisrnet}.
Both full-precision image features and Bird's Eye View (BEV) features, represented as $\textbf{X}_f \in \mathbb{R}^{C \times H \times W}$, serve as input for the full-precision activations.

In 3D occupancy networks, features transform from dense 2D space to sparse 3D space and then back to dense 3D space, causing significant differences in feature distribution.
Each module has distinct densities and distributions.

To address the problem of significant differences in feature distribution, we follow the approach of BiSRNet, employing channel-wise feature redistribution:
\begin{equation}
  \textbf{X}_r = k \cdot \textbf{X}_f + b
\end{equation}
Here, $\textbf{X}_r \in \mathbb{R}^{C \times H \times W}$ represents the activations after channel-wise feature redistribution, and $k, b \in \mathbb{R}^C$ are learnable parameters. 
$k$ represents the learnable density of redistribution, while $b$ represents the learnable bias of redistribution.

Next, $\textbf{X}_r$ is passed through the Sign function to binarize it, yielding 1-bit binarized activations $\textbf{X}_b \in \mathbb{R}^{C \times H \times W}$, as follows:
\begin{equation}
  x_b = \text{Sign}(x_r) = 
  \begin{cases} 
    +1, & \text{if } x_r > 0 \\ 
    -1, & \text{if } x_r \leq 0 
  \end{cases} \label{eq:sign}
\end{equation}
where $x_r \in \textbf{X}_r$, $x_b \in \textbf{X}_b$.

Since the Sign function is not differentiable, approximation functions are required to ensure successful backpropagation. 
Common approximation functions include piecewise linear function $\text{Clip}(\cdot)$, piecewise quadratic function $\text{Quad}(\cdot)$, and hyperbolic tangent function $\text{Tanh}(\cdot)$. We use the hyperbolic tangent function as the approximation function, defined as:
\begin{equation}
  x_b = \text{Tanh}(\alpha x_r) = \frac{e^{\alpha x_r} - e^{- \alpha x_r}}{e^{\alpha x_r} + e^{- \alpha x_r}}
\end{equation}
The Tanh function ensures gradients exist even when weights and activations exceed 1, allowing parameter updates downstream during backpropagation.

In the binarized convolutional layer, the 32-bit precision weights $\textbf{W}_f$ are binarized into 1-bit binarized weights $\textbf{W}_b$ according to the following formula:
\begin{equation}
  w_b = \mathbb{E}_{w_f \in \textbf{W}_f}(|w_f|) \cdot \text{Sign}(w_f)
\end{equation}
Here, $\mathbb{E}_{w_f \in \textbf{W}_f}(|w_f|)$ represents the average absolute value of the full-precision weights, which serves as a scaling factor to reduce the discrepancy between the binarized weights $\textbf{W}_b$ and the full-precision weights $\textbf{W}_f$. 
Multiplying this value by $\text{Sign}(w_f) = \pm 1$ yields element-wise binarized weights $w_b$.

\begin{figure}[!t]
  \centering
    \includegraphics[width=1.0\linewidth]{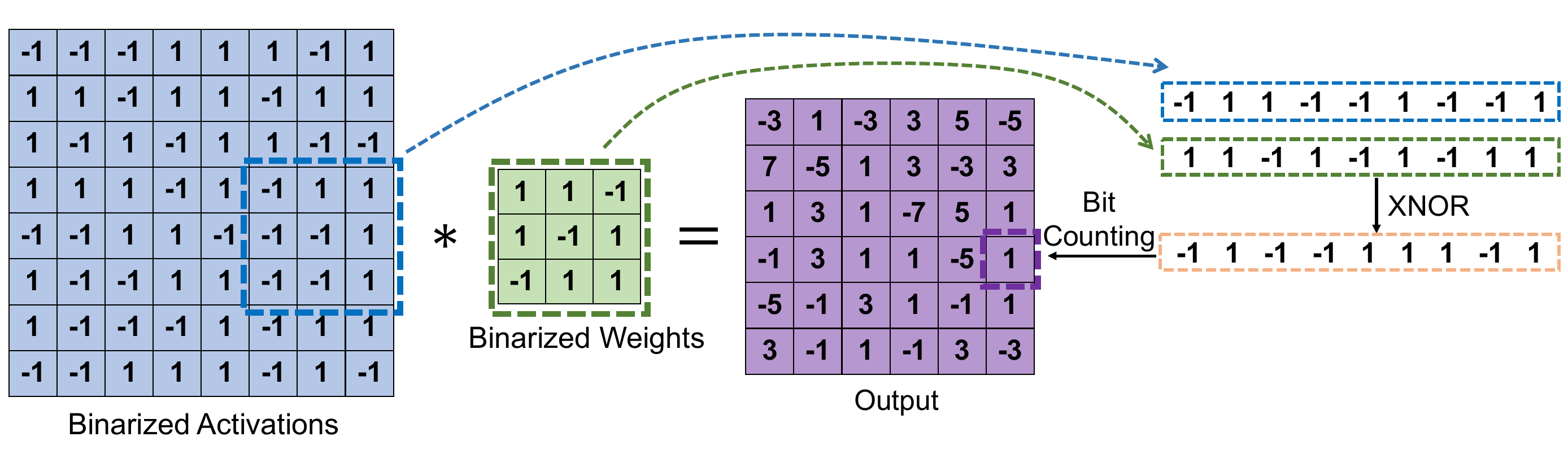}
    \caption{The schematic diagram of binarized convolution~\cite{xnornet}.
}
  \label{fig:biocc_bnn}
\end{figure}

Subsequently, the binarized activation $\textbf{X}_b$ is convolved with the binarized weights $\textbf{W}_b$.
Binarized convolution can be accomplished purely through logical operations.
The schematic diagram of binarized convolution~\cite{xnornet} is illustrated in Figure~\ref{fig:biocc_bnn}, and the expression is as follows:
\begin{equation}
  \textbf{Y}_b = \text{Biconv}(\textbf{X}_b, \textbf{W}_b) = \text{BitCount}(\text{XNOR}(\textbf{X}_b, \textbf{W}_b))
\end{equation}
Here, $\textbf{Y}_b$ is the output of binarized convolution, $\text{Biconv}$ denotes the binarized convolution layer, and $\text{BitCount}$ and $\text{XNOR}$ represent the bit count and logical XOR operations, respectively. 
In BDC-V0, the convolutional kernel size is $3 \times 3$.

For the activation function, we utilize RPReLU, whose expression is defined as follows:
\begin{equation}
  \text{RPReLU}(y_i) = 
  \begin{cases} 
    y_i - \gamma_i + \zeta_i, & \text{if } y_i > \gamma_i \\
    \beta_i \cdot (y_i - \gamma_i) + \zeta_i, & \text{if } y_i \leq \gamma_i 
  \end{cases}
\end{equation}
Here, $y_i \in \mathbb{R}$ represents the $i$-th element value of $\textbf{Y}_b$, and $\beta_i$, $\gamma_i$, and $\zeta_i$ are learnable parameters for the $i$-th channel.



\subsection{Proof of Theorem 1}

\label{sec:proof}

\textbf{Theorem 1.}
\textit{
    In the process of backpropagation, we denote the expected value of the element-wise absolute gradient error of the parameters $\textbf{w}$ in the $l$-th binarized convolutional layer as $\mathbb{E}[\Delta \frac{\partial L}{\partial w_{mn}^{(l)}}]$.
    The specific expression is as follows:
  }
  \begin{equation}
       \mathbb{E}[\Delta \frac{\partial L}{\partial w_{mn}^{(l)}}] \approx 0.5354 \cdot (\sum_i \sum_j  \sum_{m'=-(k//2)}^{k//2} \sum_{n'=-(k//2)}^{k//2} \mathbb{E}[|\frac{\partial \sigma(y_{(i+m')(j+n')}^{(l)})}{\partial y_{ij}^{(l)}} \cdot  w_{m'n'}^{(l+1)} \cdot \frac{\partial L}{\partial y_{ij}^{(l+1)}}|])
    \end{equation}
\textit{
  where $k$ is the binarized convolution kernel size, $\frac{\partial \sigma(y_{(i+m')(j+n')}^{(l)})}{\partial y_{ij}^{(l)}}$ is the derivative of the activation function $\sigma(\cdot)$, $w_{m'n'}^{(l+1)}$ represents the weights of the binarized convolutional kernel in the next layer, and $\frac{\partial L}{\partial y_{ij}^{(l+1)}}$ is the element-wise gradient in the next layer.
  }

\noindent \textbf{Proof.} 
We assume the element of the input of a binarized convolutional layer as $x_{ij}$, with a binarization error denoted as $\epsilon_{ij}$, the full-precision input before binarization as $\hat{x}_{ij}$, and the output of the binarized convolutional layer as $y_{ij}$.
Thus, we have:
\begin{equation}
  x_{ij} = \hat{x}_{ij} + \epsilon_{ij} \label{eq:error}
\end{equation}
Since the full-precision input $\hat{x}_{ij}$ at the current layer is the output from the batch normalization layer in the previous layer, we can assume that the full-precision input $\hat{x}_{ij}$ follows a Gaussian distribution $\mathcal{N}(0, 1)$.
Based on Equations \eqref{eq:sign} and \eqref{eq:error}, we can then derive the distribution of $\epsilon_{ij}$ as follows:
\begin{equation}
  |\epsilon_{ij}| = |\hat{x}_{ij} - x_{ij}| = |\hat{x}_{ij} - \text{Sign}(\hat{x}_{ij})| = 
  \begin{cases} 
    |\hat{x}_{ij} - 1|, & \text{if } \hat{x}_{ij} > 0 \\ 
    |\hat{x}_{ij} + 1|, & \text{if } \hat{x}_{ij} \leq 0 
  \end{cases} \label{eq:prob_epsilon}
\end{equation}
Assuming the convolution kernel size $k$ is odd, for a $k \times k$ convolutional layer, the kernel weight $w_{mn}$, and the kernel bias is $b_{mn}$. 
The forward propagation equation is given by:
\begin{equation}
  y_{ij} = \sum_{m=-(k//2)}^{k//2} \sum_{n=-(k//2)}^{k//2} (x_{(i+m)(j+n)} \cdot w_{mn} + b_{mn}) \label{eq:conv}
\end{equation}
Assuming that during backpropagation, the gradient at current layer $l$ is given by $\frac{\partial L}{\partial y_{ij}^{(l)}}$, we can use the chain rule to derive the gradient for a $k \times k$ convolutional layer as follows:
\begin{equation}
\begin{aligned}
    \frac{\partial L}{\partial w_{mn}^{(l)}} & = \sum_i \sum_j x_{(i+m)(j+n)}^{(l)}  \frac{\partial L}{\partial y_{ij}^{(l)}}  = \sum_i \sum_j (\hat{x}_{(i+m)(j+n)}^{(l)} + \epsilon_{(i+m)(j+n)}^{(l)}) \cdot \frac{\partial L}{\partial y_{ij}^{(l)}} 
\end{aligned} \label{eq:w_gradient}
\end{equation}
Given that the output of the current layer $y_{ij}^{(l)}$ becomes the input of the next layer after passing through the activation function $\sigma(\cdot)$.
Based on Equation \eqref{eq:conv}, we can derive:
\begin{equation}
\begin{aligned}
    y_{ij}^{(l+1)} & = \sum_{m'=-(k//2)}^{k//2} \sum_{n'=-(k//2)}^{k//2} \sigma(y_{(i+m')(j+n')}^{(l)}) \cdot  w_{m'n'}^{(l+1)} + b_{m'n'}^{(l+1)} 
\end{aligned}
\end{equation}
We can obtain the gradient relationship between $\frac{\partial L}{\partial y_{ij}^{(l)}}$ and $\frac{\partial L}{\partial y_{ij}^{(l+1)}}$:
\begin{equation}
\begin{aligned}
    \frac{\partial L}{\partial y_{ij}^{(l)}} & = \sum_{m'=-(k//2)}^{k//2} \sum_{n'=-(k//2)}^{k//2} \frac{\partial \sigma(y_{(i+m')(j+n')}^{(l)})}{\partial y_{ij}^{(l)}} \cdot  w_{m'n'}^{(l+1)} \cdot \frac{\partial L}{\partial y_{ij}^{(l+1)}}
\end{aligned} \label{eq:y_gradient}
\end{equation}
By substituting Equation \eqref{eq:y_gradient} into Equation \eqref{eq:w_gradient}, we can obtain:
\begin{equation}
\begin{aligned}
    \frac{\partial L}{\partial w_{mn}^{(l)}} &  = \sum_i \sum_j \sum_{m'} \sum_{n'} (\hat{x}_{(i+m)(j+n)}^{(l)} + \epsilon_{(i+m)(j+n)}^{(l)}) \cdot  \frac{\partial \sigma(y_{(i+m')(j+n')}^{(l)})}{\partial y_{ij}^{(l)}} \cdot  w_{m'n'}^{(l+1)} \cdot \frac{\partial L}{\partial y_{ij}^{(l+1)}}
\end{aligned}
\end{equation}
We can derive the additional gradient error $ \Delta \frac{\partial L}{\partial w_{mn}^{(l)}}$ induced by the binarization error $\epsilon$ as follows:
\begin{equation}
  \begin{aligned}
  \Delta \frac{\partial L}{\partial w_{mn}^{(l)}} & := \text{EAE}(\frac{\partial L}{\partial w_{mn}^{(l)}} - \frac{\partial L}{\partial w_{mn}^{(l)}}|_{\epsilon=0}) \\ 
    & = \sum_i \sum_j \sum_{m'} \sum_{n'} | \epsilon_{(i+m)(j+n)}^{(l)} \cdot  \frac{\partial \sigma(y_{(i+m')(j+n')}^{(l)})}{\partial y_{ij}^{(l)}} \cdot  w_{m'n'}^{(l+1)} \cdot \frac{\partial L}{\partial y_{ij}^{(l+1)}}|
  \end{aligned} \label{eq:gradient_error}
\end{equation}
Here, EAE represents the element-wise absolute error.

By utilizing Equation \eqref{eq:prob_epsilon}, we can calculate the expected value of the absolute binarization error, denoted as $\mathbb{E}[|\epsilon_{ij}|]$:
\begin{equation}
\begin{aligned}
  \mathbb{E}[|\epsilon_{ij}|] &= \int_{0}^{\infty} |\hat{x}_{ij} - 1| \frac{1}{\sqrt{2\pi}} e^{-\frac{\hat{x}_{ij}^2}{2}} \,d\hat{x}_{ij} 
  + \int_{-\infty}^{0} |\hat{x}_{ij} + 1| \frac{1}{\sqrt{2\pi}} e^{-\frac{\hat{x}_{ij}^2}{2}} \,d\hat{x}_{ij} \\
  & = 2 (\int_{0}^{1} \frac{1-\hat{x}_{ij}}{\sqrt{2\pi}} e^{-\frac
{\hat{x}_{ij}^2}{2}} \,d\hat{x}_{ij} - \int_{1}^{\infty} \frac{1-\hat{x}_{ij}}{\sqrt{2\pi}} e^{-\frac{\hat{x}_{ij}^2}{2}} \,d\hat{x}_{ij}) 
\end{aligned} \label{eq:E_epsilon}
\end{equation} 
The Gaussian error function, often abbreviated as "$erf(x)$" is defined as follows:
\begin{equation}
 erf(x) = \frac{2}{\sqrt{\pi}} \int_{0}^{x} e^{-t^2} \,dt 
\end{equation}
  Based on the definition of the Gaussian error function and the use of the substitution rule, we can compute the integral as follows:
\begin{equation}
\begin{aligned}
  \int_{0}^{x} e^{-\frac{u^2}{2}} \,du &\xlongequal{u = \sqrt{2} t} \sqrt{2} \int_{0}^{\frac{x}{\sqrt{2}}} e^{-t^2} \,dt\\
  &= \frac{\sqrt{\pi}}{\sqrt{2}} \frac{2}{\sqrt{\pi}} \int_{0}^{\frac{x}{\sqrt{2}}} e^{-t^2} \,dt\\
    &= \frac{\sqrt{\pi}}{\sqrt{2}} erf(\frac{x}{\sqrt{2}})
\end{aligned}
\end{equation}
    We can continue the computation of the integral further.
\begin{equation}
\begin{aligned}
  \int_{a}^{b} \frac{1-x}{\sqrt{2\pi}} e^{-\frac{x^2}{2}} \,dx &= \int_{a}^{b} \frac{1}{\sqrt{2\pi}} e^{-\frac{x^2}{2}} \,dx - \int_{a}^{b} \frac{x}{\sqrt{2\pi}} e^{-\frac{x^2}{2}} \,dx \\
    &= \frac{1}{\sqrt{2\pi}} (\int_{0}^{b} e^{-\frac{x^2}{2}} \,dx - \int_{0}^{a} e^{-\frac{x^2}{2}} \,dx - e^{-\frac{a^2}{2}} + e^{-\frac{b^2}{2}})\\
    &= \frac{1}{\sqrt{2\pi}} [(\frac{\sqrt{\pi}}{\sqrt{2}} erf(\frac{b}{\sqrt{2}}) - \frac{\sqrt{\pi}}{\sqrt{2}} erf(\frac{a}{\sqrt{2}}) - e^{-\frac{a^2}{2}} + e^{-\frac{b^2}{2}}]\\
\end{aligned}
\end{equation}
Equation \eqref{eq:E_epsilon} can be written as follows:
\begin{equation}
\begin{aligned}
  \mathbb{E}[|\epsilon_{ij}|] & = \frac{2}{\sqrt{2\pi}} \{[\frac{\sqrt{\pi}}{\sqrt{2}} erf(\frac{1}{\sqrt{2}}) - \frac{\sqrt{\pi}}{\sqrt{2}} erf(\frac{0}{\sqrt{2}}) - e^{-\frac{0}{2}} + e^{-\frac{1}{2}} ] \\ 
  & - [(\frac{\sqrt{\pi}}{\sqrt{2}} erf(\frac{\infty}{\sqrt{2}}) - \frac{\sqrt{\pi}}{\sqrt{2}} erf(\frac{1}{\sqrt{2}}) - e^{-\frac{1}{2}} + e^{-\frac{\infty}{2}}] \} \\
  & \xlongequal{erf(0) = 0, erf(\infty) = 1} 2[erf(\frac{1}{\sqrt{2}}) - \frac{1}{2} - \frac{1}{\sqrt{2 \pi}} + \frac{2}{\sqrt{2 \pi e}}] \approx 0.5354
\end{aligned}
\end{equation}
Therefore, based on Equations \eqref{eq:gradient_error}, the expected value of the additional gradient error $\mathbb{E}[\Delta \frac{\partial L}{\partial w_{mn}^{(l)}}]$ can be expressed  as follows:
\begin{equation}
\begin{aligned}
   \mathbb{E}[\Delta \frac{\partial L}{\partial w_{mn}^{(l)}}] &= \sum_i \sum_j \sum_{m'} \sum_{n'} \mathbb{E}[| \epsilon_{(i+m)(j+n)}^{(l)} \cdot  \frac{\partial \sigma(y_{(i+m')(j+n')}^{(l)})}{\partial y_{ij}^{(l)}} \cdot  w_{m'n'}^{(l+1)} \cdot \frac{\partial L}{\partial y_{ij}^{(l+1)}}|]
\end{aligned} \label{eq:E_gradient}
\end{equation}
Based on Equation \eqref{eq:prob_epsilon}, since the binarization error \(\epsilon_{ij}^{(l)}\) depends solely on the input \(x_{ij}^{(l)}\) and is independent of any other variables, \(\epsilon_{ij}^{(l)}\) and other random variables in Equation \eqref{eq:E_gradient} are mutually independent. 
Therefore, it follows that:
\begin{equation}
\begin{aligned}
   \mathbb{E}[\Delta \frac{\partial L}{\partial w_{mn}^{(l)}}] &= \sum_i \sum_j \sum_{m'} \sum_{n'} \mathbb{E}[|\epsilon_{(i+m)(j+n)}^{(l)}|] \cdot \mathbb{E}[|\frac{\partial \sigma(y_{(i+m')(j+n')}^{(l)})}{\partial y_{ij}^{(l)}} \cdot  w_{m'n'}^{(l+1)} \cdot \frac{\partial L}{\partial y_{ij}^{(l+1)}}|] \\
   &\approx 0.5354 \cdot (\sum_i \sum_j  \sum_{m'=-(k//2)}^{k//2} \sum_{n'=-(k//2)}^{k//2} \mathbb{E}[|\frac{\partial \sigma(y_{(i+m')(j+n')}^{(l)})}{\partial y_{ij}^{(l)}} \cdot  w_{m'n'}^{(l+1)} \cdot \frac{\partial L}{\partial y_{ij}^{(l+1)}}|])
\end{aligned} 
\end{equation}
From the above equations, it is evident that as the size \(k\) of the convolutional kernel in the subsequent layer increases, the element-wise gradient error introduced during the binarization process also increases. 
Consequently, in binarized convolutional units, the smaller the size of the convolutional kernel \(k\), the smaller the binarization error introduced into the binarized model.

Therefore, we use $1 \times 1$ binarized convolution as the new binarized convolution.

\subsection{More Details About Experiments}

\subsubsection{Result of Different Version of BDC}

We tested the performance metrics of different versions of BDC on the Occ3d-nuScenes validation set.
Table~\ref{tab:biocc_nus_sup} presents the results.
The configurations of BDC-B and BDC-T follow the settings outlined in Table \ref{tab:biocc_nus}.
We binarized all modules in the 3D occupancy network except for the view transformer, referring to this as the \textbf{small} version (\textbf{-S}).
These modules include the image encoder, the BEV encoder, and the occupancy head.

Compared to BDC-T, BDC-S additionally binarizes the image backbone in the image encoder. 
The image backbone contains substantial pre-trained knowledge, and binarizing it hinders leveraging this pre-trained knowledge, which leads to a significant performance drop compared to BDC-T.
Compared to FlashOcc$\dagger$, which does not use pre-trained weights in the image backbone, the binarized version shows a significant performance decline.

Therefore, we recommend against binarizing the image backbone.

\definecolor{nothers}{RGB}{0, 191, 255}
\definecolor{nbarrier}{RGB}{255, 120, 50}
\definecolor{nbicycle}{RGB}{255, 192, 203}
\definecolor{nbus}{RGB}{255, 255, 0}
\definecolor{ncar}{RGB}{0, 150, 245}
\definecolor{nconstruct}{RGB}{0, 255, 255}
\definecolor{nmotor}{RGB}{200, 180, 0}
\definecolor{npedestrian}{RGB}{255, 0, 0}
\definecolor{ntraffic}{RGB}{255, 240, 150}
\definecolor{ntrailer}{RGB}{135, 60, 0}
\definecolor{ntruck}{RGB}{160, 32, 240}
\definecolor{ndriveable}{RGB}{255, 0, 255}
\definecolor{nother}{RGB}{139, 137, 137}
\definecolor{nsidewalk}{RGB}{75, 0, 75}
\definecolor{nterrain}{RGB}{150, 240, 80}
\definecolor{nmanmade}{RGB}{230, 230, 250}
\definecolor{nvegetation}{RGB}{0, 175, 0}
\definecolor{lightgray}{gray}{0.9}
\definecolor{BNN}{rgb}{0.981,0.961,0.941}
\definecolor{gray}{gray}{0.85}

\definecolor{CNN}{rgb}{0.95,1,0.95}

\begin{table*}[t]

\scriptsize
 	\setlength{\tabcolsep}{0.0023\linewidth}
	
	\newcommand{\classfreq}[1]{{~\tiny(\nuscenesfreq{#1}\%)}}  %
    \begin{center}
    \caption{
    \textbf{Comparison of the occupancy prediction performance across different versions of BDC.} 
    BDC-S binarizes all modules in the 3D occupancy network except for the view transformer.
    These modules include an image encoder, BEV encoder, and occupancy head.
    $\dagger$ stands for not using pre-trained weights from an image backbone.
    }
      \vspace{-3mm} 
    \label{tab:biocc_nus_sup}

	\begin{tabular}{l|c|c|ccccccccccccccccc|c}
		\toprule
		Methods

		& \rotatebox{90}{Params(M)} & \rotatebox{90}{OPs(G)}
  
        & \rotatebox{90}{\textcolor{nothers}{$\blacksquare$} others}
        
		& \rotatebox{90}{\textcolor{nbarrier}{$\blacksquare$} barrier}
		
		& \rotatebox{90}{\textcolor{nbicycle}{$\blacksquare$} bicycle}
		
		& \rotatebox{90}{\textcolor{nbus}{$\blacksquare$} bus}

		& \rotatebox{90}{\textcolor{ncar}{$\blacksquare$} car}

		& \rotatebox{90}{\textcolor{nconstruct}{$\blacksquare$} const. veh.}

		& \rotatebox{90}{\textcolor{nmotor}{$\blacksquare$} motorcycle}

		& \rotatebox{90}{\textcolor{npedestrian}{$\blacksquare$} pedestrian}

		& \rotatebox{90}{\textcolor{ntraffic}{$\blacksquare$} traffic cone}

		& \rotatebox{90}{\textcolor{ntrailer}{$\blacksquare$} trailer}

		& \rotatebox{90}{\textcolor{ntruck}{$\blacksquare$} truck}

		& \rotatebox{90}{\textcolor{ndriveable}{$\blacksquare$} drive. suf.}

		& \rotatebox{90}{\textcolor{nother}{$\blacksquare$} other flat}

		& \rotatebox{90}{\textcolor{nsidewalk}{$\blacksquare$} sidewalk}

		& \rotatebox{90}{\textcolor{nterrain}{$\blacksquare$} terrain}
		  
		& \rotatebox{90}{\textcolor{nmanmade}{$\blacksquare$} manmade}

		& \rotatebox{90}{\textcolor{nvegetation}{$\blacksquare$} vegetation}

        & \rotatebox{90}{mIoU}

		\\
		\midrule
          \rowcolor{CNN}\multicolumn{21}{l}{\emph{\textbf{CNN-based (32 bit)}}}\\

		\rowcolor{CNN}FlashOcc~\cite{flashocc} & 44.74 & 248.57 & 9.08 & 46.32 & 17.71 & 42.70 & 50.64 & 23.72 & 20.13 & 22.34 & 24.09 & 30.26 & 37.39 & 81.68 & 40.13 & 52.34 & 56.46 & 47.69 & 40.60 & 37.84\\
        \rowcolor{CNN}FlashOcc$\dagger$~\cite{flashocc} & 44.74 & 248.57 &6.10 & 35.78 & 0.50 & 26.97 & 42.39 & 11.16 & 7.13 & 10.99 & 10.68 & 20.95 & 24.35 & 80.60 & 40.02 & 50.44 & 55.11 & 44.67 & 38.85 & 29.81 \\
  		\midrule
        \rowcolor{BNN}\multicolumn{21}{l}{\emph{\textbf{BNN-based (1 bit)}}}\\
        \rowcolor{BNN}BDC-B & 28.19 & 134.02 & 9.24 & 43.93 & 20.25 & 39.92 & 49.90 & 23.11 & 21.44 & 21.22 & 21.98 & 31.23 & 37.09 & 80.93 & 39.41 & 51.03 & 54.88 & 46.80 & 40.04 & 37.20 \\
        \rowcolor{BNN}BDC-T & 26.82 & 129.90 & 9.22 & 44.81 & 17.56 & 40.02 & 49.94 & 24.84 & 18.63 & 21.32 & 23.27 & 32.46 & 36.43 & 81.03 & 38.74 & 51.40 & 55.55 & 47.15 & 40.76 & 37.24 \\
        
        \rowcolor{BNN}BDC-S & 3.51 & 45.30 & 3.13 & 24.25 & 6.02 & 22.21 & 36.23 & 7.29 & 5.78 & 14.11 & 14.04 & 4.86 & 22.99 & 68.21 & 14.29 & 33.52 & 36.76 & 33.20 & 30.63 & 22.21 \\
		\bottomrule
	\end{tabular}
    \end{center}

\end{table*}

\subsubsection{Operations and Parameters of Binarized Module of 3D Occupancy Network}

\definecolor{title}{gray}{0.95}
\begin{table}[t]

    \small
    \begin{center}
        \setlength{\tabcolsep}{0.017\linewidth}
        \caption{Operations and parameters of binarized module of 3D occupancy network.} 
            \label{tab:ops_params}
        \begin{tabular}{l | c | c c c c c c}
        \toprule
         & Binarized & \makecell{Image\\ Backbone} & \makecell{Image\\ Neck} & \makecell{View\\ Transformer} & \makecell{BEV\\ Backbone} & \makecell{BEV\\ Neck} &\makecell{Occupancy \\ Head}\\ 

        \midrule
        \multirow{2}{*}{\centering OPs(G)} & & 88.785 & 1.377 & 0.165 & 17.724 & 102.989 &34.755 \\
         &\checkmark & 3.700 & 0.034 & - & 0.098 & 30.071 & 11.233 \\
        \midrule
        \multirow{2}{*}{\centering Params(M)} & & 23.508 & 4.155 & 0.039 & 12.394 & 6.556 & 0.869 \\
        &\checkmark & 0.162 & 0.016 & - & 0.036 & 2.975 & 0.282 \\
        \bottomrule
        \end{tabular}
    \end{center}

\end{table}

In Table~\ref{tab:ops_params}, we investigate the changes in computation (OPs) and parameters (Params) across different modules of the 3D occupancy network before and after binarization.
The image encoder consists of the image backbone and image neck, while the BEV encoder includes the BEV backbone and BEV neck.

We do not binarize the view transformer because its 32-bit full-precision parameters and computation are already sufficient. 
Additionally, the view transformer relies on full-precision computation to precisely map 2D image features to 3D BEV features.

\subsubsection{More Visualization}

\begin{figure}[t]
  \centering
    \includegraphics[width=1.0\linewidth]{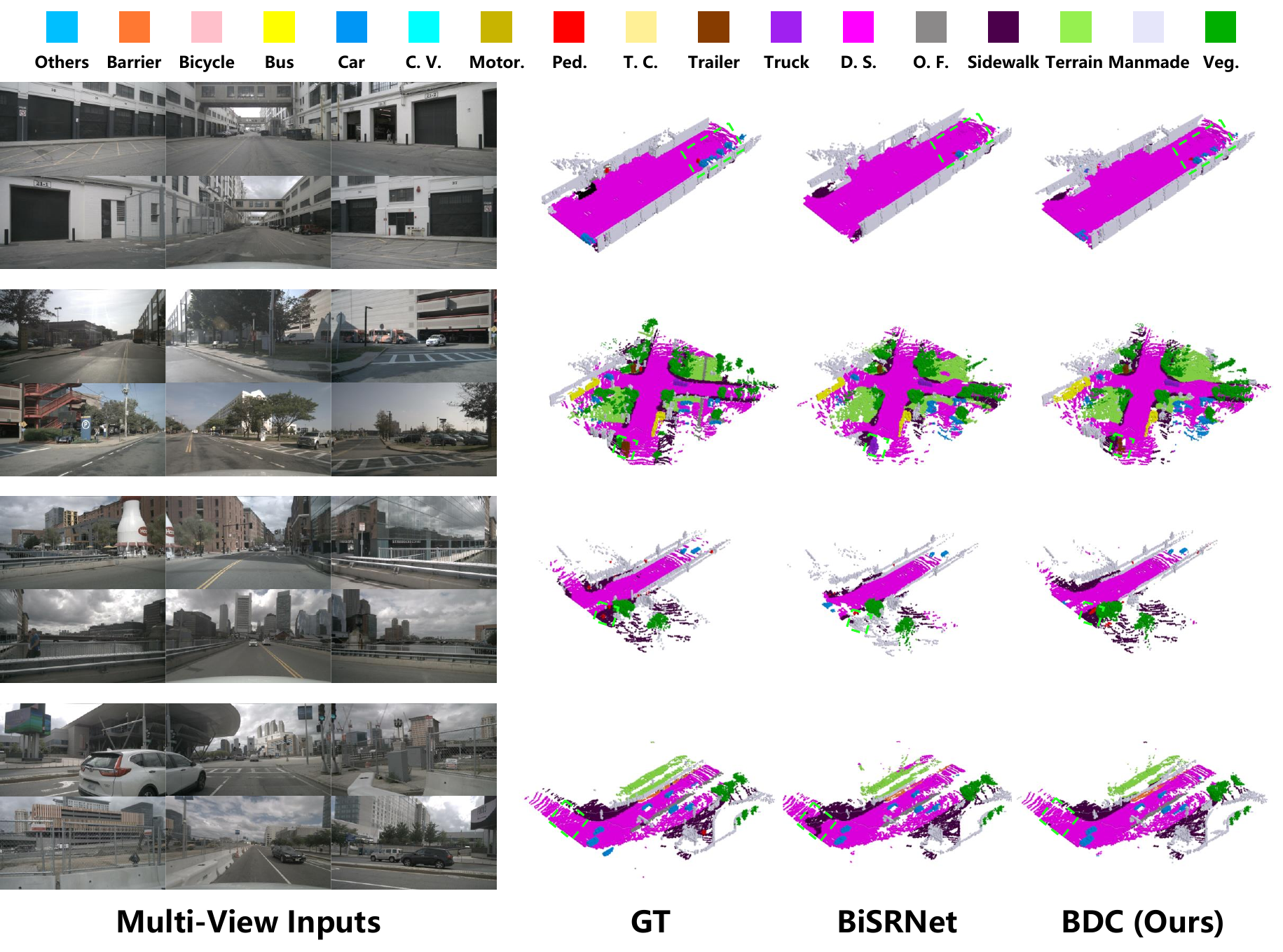}
    \caption{More Visualization rensults on Occ3D-nuScenes validation set}
  \label{fig:biocc_vis_sup}
    \vspace{-3mm} 
\end{figure}

In this section, we provide additional occupancy prediction results of BiSRNet~\cite{bisrnet} and our BDC applied to Flashocc in Fig~\ref{fig:biocc_vis_sup}.
Compared to BiSRNet, BDC offers superior scene reconstruction capability and more accurate label prediction.

\subsection{Broader Impacts}
\label{sec:broader_impacts}

3D occupancy prediction stands as a core task in autonomous driving perception. 
Leveraging occupancy grids effectively addresses real-world challenges such as long-tail datasets and target truncation, which 3D object detection algorithms may struggle to resolve.
Our approach, BDC-Occ, demonstrates superior efficiency and accuracy in predicting the occupancy status of voxels in 3D space compared to all existing state-of-the-art methods based on Binarized Neural Networks (BNNs), holding significant value for practical applications. 
Thus far, 3D occupancy prediction technology has not yielded any adverse societal impacts. 
Our proposed BDC-Occ likewise does not introduce any foreseeable negative social consequences.

\end{document}